\documentclass[authoryear,3p,preprint,12pt,authoryear]{elsarticle}

\usepackage{amssymb}
\usepackage{amsmath,amsfonts}
\usepackage{algorithm}
\usepackage{algpseudocode}
\usepackage{array}
\usepackage[caption=false,font=normalsize,labelfont=sf,textfont=sf]{subfig}
\usepackage{textcomp}
\usepackage{stfloats}
\usepackage{hyperref}
\usepackage{verbatim}
\usepackage{graphicx}
\usepackage{wrapfig}
\usepackage{xcolor}
\usepackage{tabularx}
\usepackage{lineno}
\usepackage{booktabs}
\usepackage[table]{xcolor}
\definecolor{lightgreen}{rgb}{0.9,1.0,0.9}
\usepackage{enumitem}
\setlist[itemize]{leftmargin=*}
\hypersetup{
    hidelinks,
    pdfstartview={FitH}
}
\journal{Expert Systems With Application}

\begin{document}

\begin{frontmatter}



\title{Continual Test-Time Adaptation for Object Detection with Adaptive Monitoring and Randomized Restoration} 

\author[inst1]{Shilei Cao}
\author[inst1,inst5]{Juepeng Zheng\corref{cor1}}
\author[inst2]{Yan Liu}
\author[inst1]{Baoquan Zhao}
\author[inst3]{Ziqi Yuan}
\author[inst4]{Weijia Li}
\author[inst1]{Runmin Dong}
\author[inst4,inst5,inst6]{Haohuan Fu}

\affiliation[inst1]{
            organization={School of Artificial Intelligence, Sun Yat-Sen University},
            city={Zhuhai},
            postcode={519080}, 
            country={China}}

\affiliation[inst2]{
            organization={School of Information Science and Technology, University of Science and Technology of China},
            city={Hefei},
            postcode={230026}, 
            country={China}}
            
\affiliation[inst3]{
            organization={State Key Laboratory of Intelligent Technology and Systems, Department of Computer Science and Technology, Tsinghua University},
            city={Beijing},
            postcode={100084}, 
            country={China}}

\affiliation[inst4]{
            organization={Tsinghua Shenzhen International Graduate School, Tsinghua University},
            city={Shenzhen},
            postcode={518071}, 
            country={China}}
            
\affiliation[inst5]{
            organization={National Supercomputing Center in Shenzhen},
            city={Shenzhen},
            postcode={518055}, 
            country={China}}
            
\affiliation[inst6]{
            organization={Ministry of Education Key Laboratory for Earth System Modeling and the Department of Earth System Science, Tsinghua University},
            city={Beijing},
            postcode={100084}, 
            country={China}}
            
\cortext[cor1]{Corresponding author: zhengjp8\@mail.sysu.edu.cn}

\begin{abstract}
Real-world application models are commonly deployed in dynamic environments, where the target domain distribution undergoes temporal changes.
Continual Test-Time Adaptation (CTTA) has recently emerged as a promising technique to gradually adapt a source-trained model to continually changing target domains. 
Despite recent advancements in addressing CTTA, two critical issues remain:
\textit{1)} Fixed thresholds for pseudo-labeling in existing methodologies lead to low-quality pseudo-labels, as model confidence varies across categories and domains;
\textit{2)} Stochastic parameter restoration methods for mitigating catastrophic forgetting fail to preserve critical information effectively, due to their intrinsic randomness.
To tackle these challenges for detection models in CTTA scenarios, we present AMROD, featuring three core components.
Firstly, the object-level contrastive learning module extracts object-level features for contrastive learning to refine the feature representation in the target domain.
Secondly, the adaptive monitoring module dynamically skips unnecessary adaptation and updates the category-specific threshold based on predicted confidence scores to enable efficiency and improve the quality of pseudo-labels. 
Lastly, the adaptive randomized restoration mechanism selectively reset inactive parameters with higher possibilities, ensuring the retention of essential knowledge. 
We demonstrate the effectiveness of AMROD on four CTTA object detection tasks, where AMROD outperforms existing methods, especially achieving a 3.2 mAP improvement and a 20\% increase in efficiency on the Cityscapes-to-Cityscapes-C CTTA task. 
The code of this work is available at \href{https://github.com/ShileiCao/AMROD}{https://github.com/ShileiCao/AMROD}.
\end{abstract}

\begin{keyword}
Unsupervised Domain Adaptation \sep Test-Time Adaptation \sep Continual Test-Time Adaptation \sep Object Detection
\end{keyword}

\end{frontmatter}


\section{Introduction}
Deep learning models have demonstrated immense potential across various vision tasks
such as image recognition \citep{he2016deep}, object detection \citep{ren2015faster}, and image segmentation \citep{long2015fully}. 
However, these models experience pronounced degradation in performance when confronted with training data (i.e., source domain) and testing data (i.e., target domain) originating from disparate distributions. 
This phenomenon, commonly referred to as distribution shifts, poses a significant challenge \citep{hendrycks2019benchmarking}. 
In such a scenario, unsupervised domain adaptation (UDA) becomes crucial, which typically involves aligning the distributions of source and target data, thereby mitigating the impact of distribution shifts \citep{zhang2022multi,jiang2025multi}.
Still, UDA falls short by necessitating access to source data, which is often inaccessible due to privacy constraints, proprietary data concerns, or data transmission barriers \citep{vs2023towards}.
This limitation catalyzes the exploration of Test-Time Adaptation (TTA) where the source-trained model directly adapts toward unlabeled test samples encountered during evaluation in an online manner, without the reliance on the source data \citep{wang2020tent}. 
Nonetheless, the TTA methods, which assume a static target domain, face a more challenging and realistic problem, as real-world systems work in non-stationary environments. 
For example, a vehicle may encounter various continuous environmental changes such as fog, night, rain, and snow during its journey.
Existing TTA methods are vulnerable to catastrophic forgetting of source knowledge and error accumulation when adaptation faces more than one distribution shift \citep{wang2022continual}.
These issues become more pronounced when adaptation occurs in environments where the target domain is not only dynamic but also evolves over time.

\begin{figure}[tbp]
    \centering
    \includegraphics[width=\textwidth]{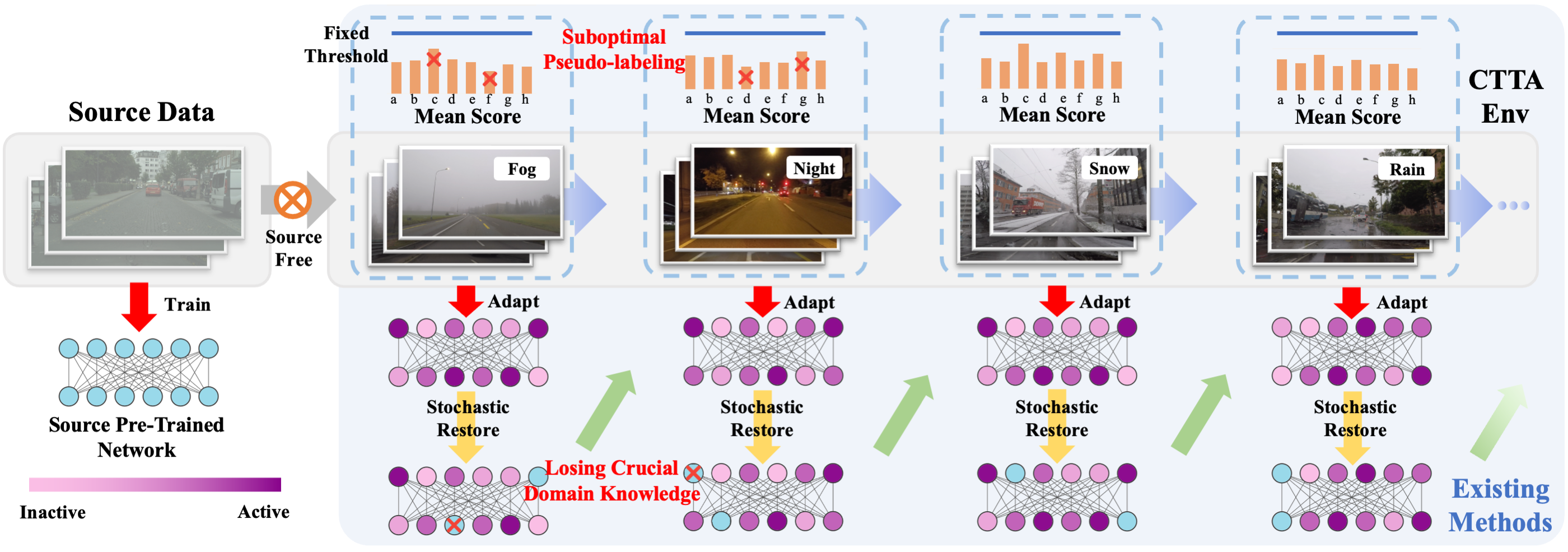}
    
    \caption{Motivation.
    {(1) Top: Model confidence fluctuates across different categories and target domains. 
    Applying a uniformly fixed threshold fails to capture these variations, leading to suboptimal pseudo-labeling. 
    (2) Bottom: Stochastic restoration randomly resets neurons to mitigate forgetting (darker colors indicate higher activity, while blue represents source knowledge). 
    This randomness can inadvertently erase crucial, domain-specific knowledge acquired from adaptation.}
    }
    \label{challenge}
    
\end{figure}

Recently, \citet{wang2022continual} introduce CoTTA by applying stochastic parameters restoration to mitigate catastrophic forgetting in Continual Test-Time Adaptation (CTTA) scenarios, where the model is continually adapted to sequences of target domains. 
Although randomly resetting parameters partially helps alleviate the forgetting of source knowledge \citep{wang2022continual}, its randomness may also contribute to losing crucial knowledge specific to the current domain (refer to Figure \ref{challenge}).
Additionally, existing CTTA methods normally utilize pseudo-labeling through a fixed threshold for self-supervision 
\citep{wang2022continual,dobler2023robust}.
Given that model confidence may vary across categories and domains (refer to Figure \ref{challenge}), employing a uniformly fixed threshold could exclude high-quality pseudo-labels while incorporating incorrect ones.
Furthermore, these methods adapt to
every new incoming data, resulting in computational inefficiency and potential performance degradation.
Low-quality pseudo-labels and unstable adaptation to noisy data lead to error accumulation, as they provide negative feedback to the model, undermining its performance.

{Therefore}, two significant {challenges} persist in CTTA: 
Current self-training-based methods suffer from low-quality pseudo-labels, leading to error accumulation (\textbf{\textit{Challenge 1}}); 
Continual adaptation to dynamic environments struggles to effectively retain valuable knowledge about current domain while mitigating forgetting of source knowledge (\textbf{\textit{Challenge 2}}).

To tackle these challenges, we propose AMROD (\textbf{A}daptive \textbf{M}onitoring and \textbf{R}estoration for \textbf{O}bject \textbf{D}etection). 
Aligning with previous CTTA works for robust adaptation \citep{wang2022continual,dobler2023robust}, {AMROD} is constructed upon the mean-teacher framework \citep{tarvainen2017mean}.
This framework involves a student model supervised by a teacher model,
where the teacher model is an exponential moving average of the student model. 
Particularly, {AMROD} comprises Object-level Contrastive Learning (OCL), Adaptive Monitoring (AM), and Adaptive Randomized Restoration (ARR) modules.

More specifically, 
OCL extracts object-level features based on Region Proposal Network (RPN) proposals, which provide multiple cropped views around the object at different locations and scales.
Subsequently, Contrastive Learning (CL) loss is applied on the proposals to guide the model to encourage similar object instances to remain close while pushing dissimilar ones apart. 
The OCL is well integrated into the mean-teacher paradigm as a drop-in enhancement to acquire more fine-grained feature representation.
Furthermore, to make adaptation more efficient and stable and to improve the quality of pseudo-labels, i.e., addressing \textbf{\textit{Challenge 1}}, we design an AM module to decide whether to pause or resume the adaptation and dynamically adjust the category-specific thresholds, based on the mean predicted confidence scores.
The dynamic nature of the AM method makes it better suited to address the effects of continuously changing distributions.
Finally, the proposed ARR mechanism resets inactive parameters with a higher possibility than active ones, by utilizing Fisher information \citep{fisher1922mathematical} as an indicator of parameter importance while incorporating a stochastic component.
ARR not only helps prevent forgetting but preserves important information, i.e., addressing \textbf{\textit{Challenge 2}}. 
On the other hand, its randomness allows falsely activated parameters to be reset as well, thereby resulting in greater stability for adaptation.

We demonstrate the effectiveness of AMROD on four CTTA object detection benchmarks, which involve synthetic and real-world distribution shifts in the short- and long-term adaptation, i.e. 
Cityscapes \citep{cordts2016cityscapes}, Cityscapes-C \citep{hendrycks2019benchmarking}, SHIFT \citep{sun2022shift}, and ACDC \citep{sakaridis2021acdc} datasets. 
Results indicate that our method significantly improves performance over existing methods, with gains of up to 3.2 mAP and 20\% in computational efficiency. Our main contributions are:
\begin{itemize}
    \item This study introduces {AMROD}, which pioneers in exploring CTTA for detection models. Specifically, we propose to leverage object-level features for contrastive learning to refine feature representation in CTTA object detection{, bypassing the computational burden of large batch sizes typically required for contrastive learning.} 
    \item To address the two challenges in CTTA, our proposed {AM module enables dynamic skipping and category-specific threshold updates based on the mean predicted scores, therefore improving robustness and pseudo-label quality. 
    Moreover, the ARR module resets the inactive parameter with higher possibilities, effectively preventing error accumulation and catastrophic forgetting.}
    \item Empirical experiments demonstrate that {AMROD} surpasses existing methods and facilitates short-term and long-term adaptation {under both synthetic and real-world continual distribution shift, notably achieving up to a 3.2 mAP performance gain and a 20\% increase in computational efficiency on the Cityscapes-to-Cityscapes-C task.
    }
\end{itemize}


\section{Related works}
\label{s2}
\subsection{Source-Free Domain Adaptation}
UDA tackles the inter-domain divergence by aligning the distributions of source and target data \citep{saito2018maximum,saito2019strong,li2022cross,li2025uncertainty,xu2025unsupervised,liang2025federated,cao2025crossearth,liang2025tta}.
Despite its effectiveness, the limitation of UDA lies in its requirement for access to the source domain data, which often raises concerns regarding data privacy and transmission efficiency \citep{vs2023towards,huang2021model}.
As a result, Source-Free Domain Adaptation (SFDA) received extensive research attention, where the source-trained detector is adapted to the target training data without any source data
\citep{vs2023towards,vs2023instance,li2022source,chen2023exploiting,lu2024consistency,wang2025dit,ye2025quantifying} before evaluation.
For instance, MemCLR \citep{vs2023instance} employs a cross-attention-based memory bank with CL for source-free detectors, while IRG \citep{vs2023instance} utilizes the object relations with instance relation graph network to explore the SFDA setting for object detection.
However, the standard SFDA setting requires prior knowledge of the target domain, which is impractical in most real-world applications. 

\subsection{Test-Time Adaptation}
TTA adapts the source-trained model to the target test data during inference time without access to the source data. 
Since both TTA and SFDA involve adapting the source-trained model to the unlabeled target data without utilizing source data, some works also refer to TTA as SFDA \citep{wang2022continual,brahma2023probabilistic}.
In this paper, we distinguish between TTA and SFDA based on evaluation protocol, although they can be transformed into each other in experiments.
Furthermore, TTA methods improve the model performance under distribution shift commonly through pseudo-labeling \citep{sun2020test,iwasawa2021test,zeng2023exploring,liang2025tta}, batchnorm statistics updating \citep{hu2021mixnorm,you2021test}, or entropy regularization \citep{wang2020tent,iwasawa2021test,niu2022efficient,fu2025robust,liang2025low} during testing.
For example, Tent \citep{wang2020tent} updates the batchnorm parameters with entropy minimization and demands a large batch size for optimization during test-time adaptation, which is unsuitable for real-time detection model deployment where images are processed sequentially.
The above approaches assume a static target domain where the target data come from a single domain.
However, in practical scenarios, the distribution of target domain may exhibit a continual shift over time.

\subsection{Continual Test-time Adaptation}
{
Conceptually, the core principles of CTTA are closely related to the classical paradigm of on-line retrainable neural networks \citep{doulamis2000line,ioannou2006adaptive,an2011neural,kollias2016line}. 
Originally explored in dynamic vision and signal processing tasks such as video segmentation and continuous image analysis \citep{doulamis2000line,ntalianis2002unsupervised}, retrainable neural networks are designed to continuously update their parameters during the operational (inference) phase to adapt to non-stationary environments.
A critical challenge in these early dynamic systems, as in modern CTTA, is the forgetting issues. 
To prevent the loss of previously acquired knowledge while adapting to new conditions, these historical structures explicitly handle the forgetting issue through constrained optimization and memory management. For instance, systems employ retraining algorithms that explicitly minimize weight modifications relative to previous model states \citep{doulamis2000line,doulamis2002recursive} and utilize selective retraining frameworks that actively maintain representative historical samples alongside new observations \citep{an2011neural}.

Modern CTTA formalizes this continuous adaptation challenge under unsupervised constraints. 
Early modern works} 
consider adaptation to evolving and continually changing domains by aligning the source and target data \citep{hoffman2014continuous}.
These methods rely on source data during inference, which limits their applicability. 
Recently, \citet{wang2022continual} introduce CoTTA, marking the first work tailored to the demands of CTTA by adapting a pre-trained model to sequences of domains without source data.
Subsequently, research efforts have been dedicated to exploring CTTA, primarily focused on classification \citep{dobler2023robust,niloy2024effective,brahma2023probabilistic,gan2023decorate,yu2023noise} and segmentation \citep{song2023ecotta,ni2023distribution,liu2023vida,niloy2024effective,zhu2023uncertainty} tasks.
For instance, similar to CoTTA, \citet{zhu2023uncertainty} apply a stochastic reset mechanism to prevent forgetting in the CTTA medical segmentation task. 
In contrast, PETAL \citep{brahma2023probabilistic} utilizes the Fisher Information Matrix (FIM) as a metric of parameter importance to reset only the most irrelevant parameters across all layers. 
Nevertheless, these two methods are sub-optimal, as the randomness might lead to losing essential information, and pure data-driven restoration may retain false active parameters resulting from noise. 

Moreover, the mean-teacher framework \citep{tarvainen2017mean} serves as a base architecture for most CTTA works \citep{wang2022continual,gong2022note,dobler2023robust,niloy2024effective}, where the teacher model generates pseudo-labels via a fixed threshold to supervise the training of the student model.
Nonetheless, these methods suffer from low-quality pseudo-labels with a uniform threshold since the model confidence varies across categories and domains. 
Furthermore, \citet{wang2024continual} design a dynamic thresholding technique to update the threshold in a batch for classification tasks, requiring a large batch size. 
However, this approach is unsuitable for object detection where smaller batch sizes are preferred for computational efficiency.
{For object detection,} \citet{gan2023cloud} present a cloud-device collaborative continual adaptation paradigm by aligning source and target distribution.
{More recently, \citet{yoo2024and} explore primarily on short-term CTTA by utilizing feature distribution statistics calculated from source samples to determine whether to update a lightweight adapter integrated into the backbone.}
Nevertheless, the reliance on {accessing} source data in both methods restricts their practical applicability
{due to privacy and resource limitations.
In contrast, AMROD operates under a strictly source-free constraint without relying on any source data or architectural additions, addressing forgetting for both short-term and long-term CTTA.}
Therefore, a gap still exists in exploring the CTTA in object detection to improve the quality of pseudo-label (Challenge 1) and effectively reset noisy neurons (Challenge 2), without relying on source data.

\subsection{Continual Learning}
Continual learning, also known as incremental learning, typically involves enabling the model to retain previously acquired knowledge while learning from sequences of tasks \citep{de2021continual}.
It is commonly categorized into replay methods \citep{rebuffi2017icarl,tiwari2022gcr}, parameter isolation method \citep{aljundi2017expert,xu2018reinforced}, and regularization-based methods \citep{kirkpatrick2017overcoming,li2017learning}.
As an illustration, Elastic weight consolidation (EWC) \citep{kirkpatrick2017overcoming} is a regularization-based technique that penalizes the changing of parameters with a significant impact on prediction, based on the Fisher Information Matrix (FIM).
In this paper, motivated by \citep{kirkpatrick2017overcoming,brahma2023probabilistic}, we adopt the FIM as a metric of parameter importance for resetting noisy parameters.
Furthermore, we introduce randomness to enhance model robustness during continual adaptation.
Additionally, while the continual learning approaches aim to tackle catastrophic forgetting in sequences of new tasks, our work focuses on learning from different domains for a single task.

{
\subsection{Knowledge Distillation}
The concept of Knowledge Distillation (KD) originally emerged as a model compression technique designed to transfer learned knowledge from a cumbersome, complex teacher model into a more compact student model \citep{bucilua2006model,hinton2015distilling}. 
In the realm of object detection, KD has been adapted to handle complex localization and regression tasks, effectively transferring fine-grained, region-based feature representations and relational knowledge between domains \citep{chen2017learning,tian2021knowledge}.
Beyond model compression, recent studies demonstrate that distilling knowledge from an adapted or stabilized teacher network can effectively mitigate catastrophic forgetting in exemplar-free continual learning scenarios \citep{szatkowski2024adapt} and facilitate domain-aware continual generalization \citep{reddy2024towards}.
}

{
This concept is highly relevant to our proposed AMROD and most CTTA frameworks, which utilize a mean-teacher architecture \citep{tarvainen2017mean}. 
The mean-teacher paradigm functions as an online, self-knowledge distillation mechanism, which relies on the slowly updating teacher model to distill historically stabilized knowledge into the dynamically updating student model via pseudo-labels. 
This distillation process acts as an anchor against continuous domain shifts, complementing our ARR module to ensure that the network acquires new target-domain representations without catastrophically forgetting its foundational object detection capabilities.
}

{
\subsection{Position of the Proposed Work}
While existing methods tackle isolated aspects of domain adaptation, they fall short in fully source-free, continually changing environments for object detection. 
Unlike standard UDA or SFDA methods that assume a static target domain, AMROD is explicitly designed to handle continual, non-stationary distribution shifts. 
Furthermore, unlike modern CTTA approaches for object detection that still rely heavily on source data, AMROD positions itself as a strictly standalone, fully source-free framework. 

AMROD effectively bridges the gaps between TTA, Continual Learning, and Knowledge Distillation. 
By utilizing the mean-teacher paradigm as an online self-knowledge distillation anchor, AMROD stabilizes the adaptation process. 
It directly addresses the catastrophic forgetting inherent in CTTA through the ARR module, while simultaneously overcoming the noisy pseudo-labeling problem of traditional TTA through the AM model. 
Ultimately, this positions AMROD as a highly efficient, robust solution uniquely tailored for the complex demands of real-world continual object detection.
}

\begin{table}[tbp]
\centering
\caption{Comparisons between different problem settings.}
\resizebox{\textwidth}{!}{
\setlength{\tabcolsep}{4pt}
\begin{tabular}{l|cccccc}
\toprule
Setting & Source Data & Target Training Data  & Target Distribution & Train  Loss & Test Loss & Online \\ \midrule
Continual Learning              & $\times$ & $(x^t, y^t)$ & Dynamic & $\mathcal{L}(x^t, y^t)$ & $\times$ & $\times$ \\
Unsupervised Domain Adaptation  & $(x^s, y^s)$ & $(x^t)$  & Static & $\mathcal{L}(x^s, y^s)+\mathcal{L}(x^s, x^t)$ & $\times$  & $\times$ \\
Source-free  Domain Adaptation  & $\times$ & $(x^t)$  & Static & $\mathcal{L}(x^t)$ & $\times$  & $\times$  \\
Test-time Adaptation            & $\times$ & $\times$  & Static & $\times$ & $\mathcal{L}(x^t)$  & $\checkmark$  \\
Continual Test-Time Adaptation  & $\times$ & $\times$ & Dynamic & $\times$ & $\mathcal{L}(x^t)$ & $\checkmark$ \\ \bottomrule
\end{tabular}
}
\label{t1}
\end{table}

\section{Method}
\label{s3}
\subsection{Preliminary}
\subsubsection{{Problem Statement}}

Given a sequence of domain $D = \{d_{i}\}_{i=0}^n$, we define $d_0$ as the source domain and the subsequent domains as the target domain.
The objective of CTTA is to enhance the performance of the model $M_{\theta^{0}}(x)$, where the parameters $\theta^{0}$ are pre-trained on source data $x_{d_0},y_{d_0}$ from $d_{0}$, in a continually changing target domain during inference time without using source data. 
For simplicity, we hereafter denote the data without the subscript about the specific domain.
At time step $t$, unlabeled target data $x^t$ is provided sequentially, following the domain group order. 
The model is required to make a prediction ${y}^t = M_{\theta^{t-1}}(x^t)$ using the parameters $\theta^{t-1}$ which have been updated based on previous target data $x^1, ..., x^{t-1}$. 
Subsequently, ${y}^t$ serves as the evaluation output at time step t, and the model will adapt itself toward $x^t$ as $ \theta^{t}$, which will only influence future inputs $x^{t+n}$.

Moreover, we compare CTTA with other settings in Table \ref{t1}. 
These settings are developed to meet the diverse prerequisites and requirements for real-world applications.

\subsubsection{{Mean-Teacher Framework}}
Following previous CTTA works \citep{wang2022continual}, we build AMROD based on the mean-teacher framework \citep{tarvainen2017mean}, which features the interplay between a teacher model and a student model.
Specifically, both networks are first initialized with the source-trained model.
The teacher model produces the pseudo-labels $\hat{y}^t$ based on teacher prediction $y^t$ for the unlabeled target data to supervise the student model.
While the parameters of the student are optimized via gradient descent, the teacher is updated following an Exponential Moving Average (EMA) strategy based on the student. 
Formally, this process can be expressed as follows:
\begin{eqnarray}
\label{pl}
&&\mathcal{L}_{pl}(x^t) = \mathcal{L}_{rpn}(x^t,\hat{y}^t)+\mathcal{L}_{rcnn}(x^t,\hat{y}^t), \\
\label{stu}
&&\theta^{t}_{S}\leftarrow\theta^{t-1}_{S}+\gamma\frac{\partial(\mathcal{L}_{pl}(x^t))}{\partial\theta^{t-1}_{S}},\\
\label{T}
&&\theta^{t}_{T}\leftarrow\alpha\theta^{t-1}_{T}+(1-\alpha)\theta^{t}_{S},
\end{eqnarray}
\noindent where $x^t$ and $\hat{y}^t$ denote the unlabeled target data and corresponding pseudo-labels at time step $t$. $\theta_{S}^t$ and $\theta_{T}^t$ symbolize the parameters of student and teacher networks. 
Moreover, the supervision loss $\mathcal{L}_{pl}$ in Faster-RCNN \citep{ren2015faster}, which consists of the RPN loss $\mathcal{L}_{rpn}$ and RCNN loss $\mathcal{L}_{rcnn}$, is utilized for pseudo-labeling.
Additionally, $\gamma$ represents the student's learning rate, and the EMA rate is denoted by $\alpha$.
Hence, the teacher can be regarded as an ensemble of historical students, providing stable supervision.
However, this framework encounters error accumulation and catastrophic forgetting in dynamic environments.
Therefore, we design AMROD to fit in the CTTA setting.

\begin{figure}[tbp]
    \centering
    \includegraphics[width=\textwidth]{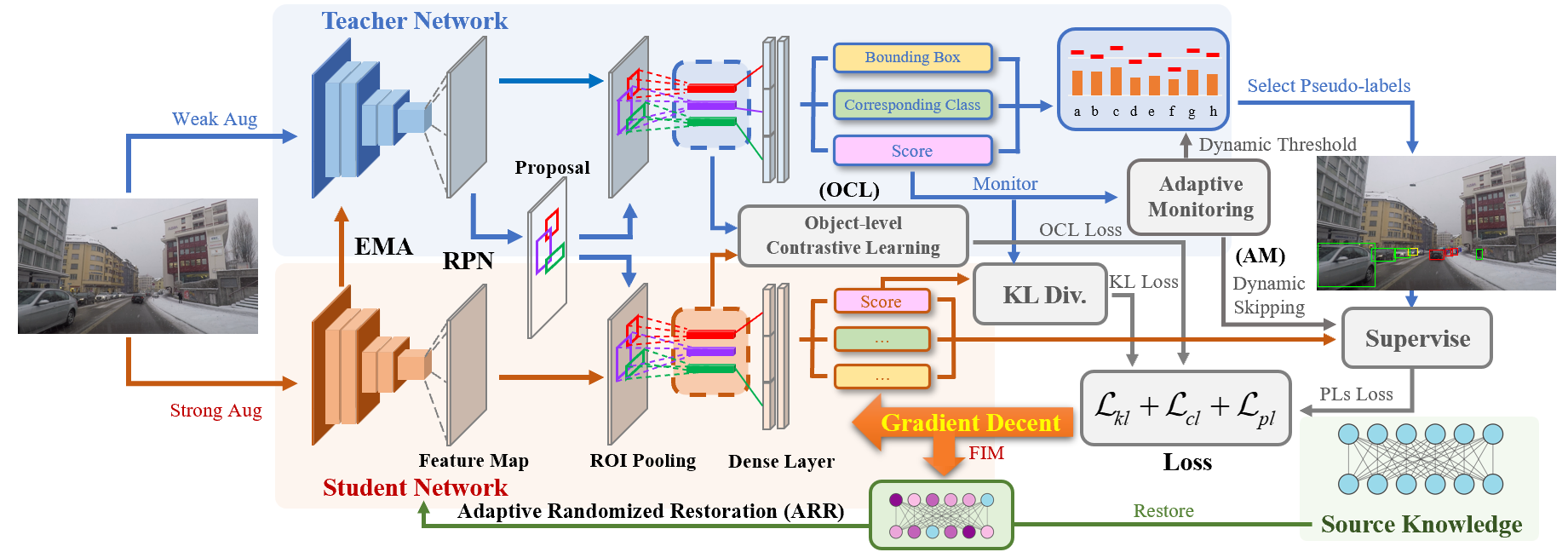}
    
    \caption{The overview of the proposed method. AMROD follows the mean-teacher framework, featuring:
    \textit{(1)} The OCL module compares region of interest features extracted from feature maps of both networks based on teacher proposals for contrastive learning; 
    \textit{(2)} The AM module dynamically skips unnecessary adaptation and adjusts category-specific thresholds based on the mean prediction scores;
    \textit{(3)} The ARR strategy reset the inactive parameter with higher possibilities based on the FIM.
    }
    
    \label{model}
\end{figure}

\subsection{Proposed Method}
\subsubsection{{Overview}}
Figure \ref{model} presents an overview of AMROD with details described in Algorithm \ref{pseudo}, consisting of Object-level Contrastive Learning (OCL), Adaptive Monitoring (AM), and Adaptive Randomized Restoration (ARR). 
Inspired by UDA object detection methods \citep{li2022cross,liu2021unbiased}, we adopt the Weak-Strong augmentation to enable the teacher model to generate reliable pseudo-labels without being affected by heavy augmentation.
Specifically, the teacher receives input with weak augmentations, while we input student networks with strong enhanced images.
The overall loss of the student is defined as:
\begin{equation}
    \mathcal{L}_{all} = \mathcal{L}_{pl}(x^t) +  \lambda \mathcal{L}_{cl}(x^t) + \mu \mathcal{L}_{kl}(x^t), 
\end{equation}
\noindent where $\mathcal{L}_{cl}(x^t)$ denotes the contrastive learning (CL) loss, to be elaborated later. The $\mathcal{L}_{kl}$ represents the Kullback-Leibler (KL) Divergence \citep{kullback1951information} loss, used to quantify the distinction between two probability distributions.
The parameters $\lambda$ and $\mu$ are corresponding weights. KL loss is defined as: 
 \begin{equation}
     \mathcal{L}_{kl}\left(P\parallel Q\right)=\sum_{x\in\mathcal{X}}P(x)\mathrm{~}\log\left(\frac{P(x)}{Q(x)}\right),
 \end{equation}
where $P(x)$ and $Q(x)$ represent different distribution. 
We employ the KL divergence loss to encourage the student model to approximate the teacher model closely.

\begin{algorithm}[t]
\centering
\small
\caption{Pseudo-code for the {AMROD}}
\label{pseudo}
\begin{algorithmic}[1]

\State \textbf{Input:} Unlabeled test data $x^t$ for the target domain
\For{each iteration $t$}
    \State Generate predictions $y^t$ using the teacher model
    \State \textcolor{gray}{//  1. Adaptive Monitoring}
    
    \State Compute the index $\frac{\overline{l}^t}{\overline{l}^{t-1}_{ema}}$
    based on teacher prediction
    \State Udapate the moving average $\overline{l}^t_{ema}$ through Eq. (\ref{amupdate})
    \If {$\frac{\overline{l}^t}{\overline{l}^{t-1}_{ema}}$ not lies with $(\frac{1}{\delta_s},\delta_s)$}
        \State {Pause the current adaption}
    \EndIf
    \State \textbf{for} each category $c$ \textbf{do}
        \State \hspace{\algorithmicindent} Compute mean predicted scores $\overline{l^t_c}$ of category $c$
        
        \State \hspace{\algorithmicindent} Update dynamic thresholds through Eq. (\ref{threshupdate})
    \State \textbf{end for}
    
    \State \textcolor{gray}{//  2. Object-level Contrastive Learning}
    
    \State Extract teacher features $T^t$ and student features $S^t$ based on teacher proposal
    \State Compute contrastive learning loss through Eq. (\ref{pl})

    \State Generate the pseudo-label $\hat{y}^t$ through $\delta^t_c$

    \State Update the student model $\theta^t_{S}$ through Eq. (\ref{stu}) with the supervised loss through Eq. (\ref{contrast}) and teacher model $\theta^t_{T}$ through Eq. (\ref{T})

    \State \textcolor{gray}{//  3. Adaptive Randomized Restoration}
    
    \State Generate random matrix $R^t \sim \operatorname{Uniform}(0,1)$
    \State Generate the FIM $F^t$ through Eq. (\ref{fim})
    \State Generate reset score matrix $W^t$ through Eq. (\ref{score})
    \State Find the $q$-quantile of $W^t$: $\eta = quantile(F^t, q)$
    \State Generate the mask matrix $M^t$ through Eq. (\ref{mask})
    \State Reset the updated student model through Eq. (\ref{weight})
    
\EndFor
\State \textbf{Output:} Teacher predictions $y^t$
\end{algorithmic}
\end{algorithm}

\subsubsection{{Object-Level Contrastive Learning}}
\label{ocl}
SimCLR \citep{chen2020simple} is a widely used CL approach for self-supervised learning, which learns high-quality feature representation across differently augmented views of the same image.
Notably, the SimCLR is initially designed for classification tasks, assuming each image pertains to a single category, and requires large batch sizes to ensure sufficient positive and negative pairs for representation learning.
Consequently, the original SimCLR is not well-suited for object detection, where images typically contain multiple instances, thus requiring significant computational resources to accommodate large batch sizes.


Motivated by SimCLR \citep{chen2020simple}, we present an OCL module to extract teacher and student features for CL based on proposals from the RPN.
This strategy provides multiple cropped views around the object instance, eliminating the need for large batch sizes and ensuring computational efficiency for online detector updates.
Specifically, given a weakly augmented image $A_{weak}(x^t)$ at time step t, the teacher produces $l$ ROI proposals $P^t = \{p_i^t\}_{i=1}^{l}$ via region proposal network. 
OCL then apply RoIAlign \citep{he2017mask} to extract corresponding teacher and student object-level features $T^t = \{ t_i^t \in\mathbb{R}^{1\times C}\}_{i=1}^{l}$ and $S^t = \{ s_i^T \in\mathbb{R}^{1\times C}\}_{i=1}^{l}$ based on the feature map from the backbone, respectively.
The features associated with the same proposal are considered positive pairs, otherwise negative pairs.
Then CL loss is applied to these features $T^t$ and $S^t$ images by minimizing:
\begin{equation}
    \mathcal{L}_{cl}(x^t) = \frac{1}{l}\sum_{i=1}^{l}-\log\frac{\exp(t_i^t\cdot s_i^t/\tau)}{\sum_{j=1}^{l}\exp(t_i^t \cdot s_j^t /\tau)},
    \label{contrast}
\end{equation}
where $\tau > 0$ is the temperature, and $t_i^t$ and $s_i^t$ denote the features of two different augmentations of the same object, serving as the positive pair.
This strategy encourages the model to learn fine-grained and localized feature representations on the target domain, without relying on accurate pseudo-labels. 
Moreover, the OCL is well integrated into the mean-teacher self-training paradigm as a drop-in enhancement for feature adaptation.

\subsubsection{{Adaptive Monitoring}}
The AM module monitors the model’s status using the predicted scores with dynamic skipping and dynamic thresholds. 
The former halts unnecessary adaptations to save computational resources, while the latter updates the category-specific threshold dynamically. 

Specifically, the teacher first make predictions ${y}^t = M_{\theta_T^{t-1}}(A_{weak}(x^t))$ for the weakly enhanced images. 
AM then computes the overall mean scores across all predicted instances  $\overline{l}^t$ in teacher's predictions and maintains its exponentially moving average $\overline{l}^t_{ema}$ through: 
\begin{equation}
\overline{l}^{t}_{ema} \leftarrow \beta_s \cdot \overline{l}^{t-1}_{ema} + (1-\beta_s) \cdot \overline{l}^t,
\label{amupdate}
\end{equation}
where $\beta_s$ represents the update rate. 
Adaptation resumes when the $\frac{\overline{l}^t}{\overline{l}^{t-1}_{ema}}$ lies within the range $(\frac{1}{\delta_s},\delta_s)$, indicating stability, and pauses otherwise, suggesting changes in data characteristics.
This dynamic skipping strategy effectively prevents unstable and noisy adaptation, thus conserving computational resources and mitigating error accumulation.

\begin{figure}[t]
    \centering
    \includegraphics[width=0.9\linewidth]{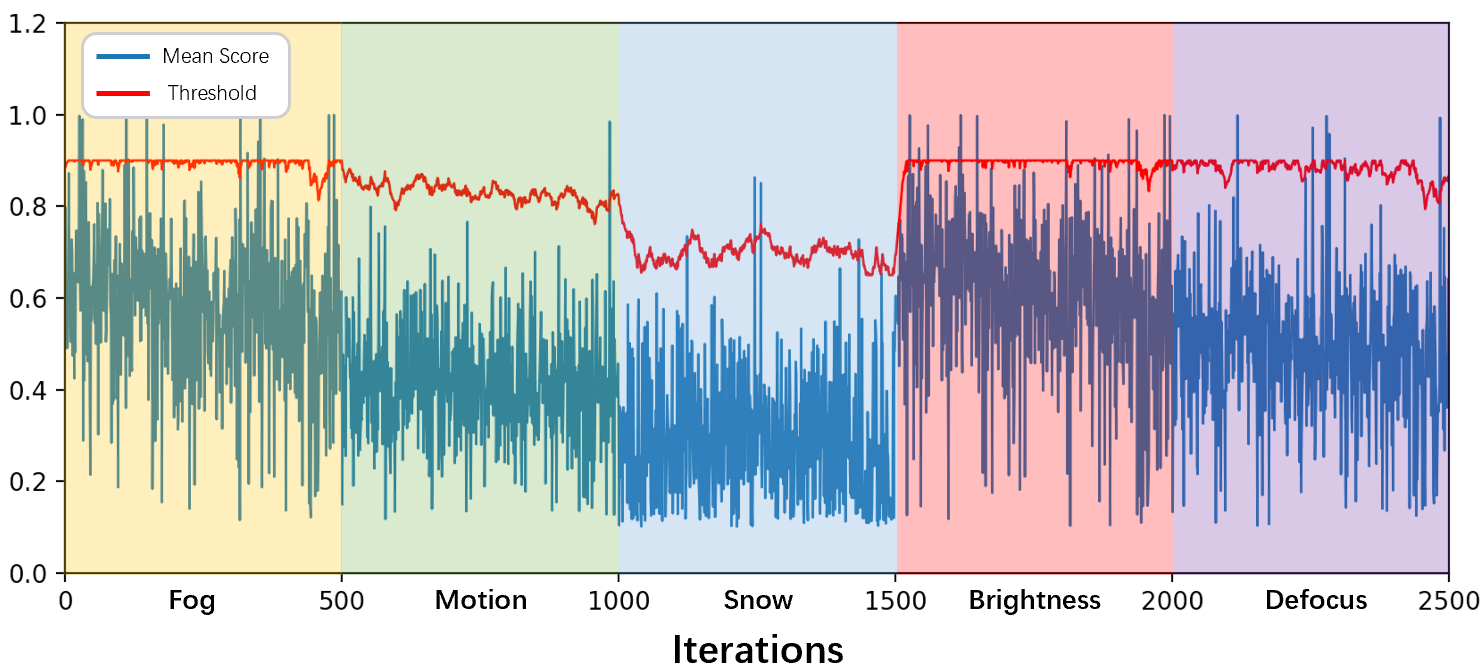}
    \caption{ Visualization of the mean score and the dynamic threshold for the "car" category during adaptation across five different domains from the Cityscapes-C dataset. The threshold (red) dynamically adjusts to the model's mean confidence score (blue) as the environment changes.}
    \label{fig:dt}
\end{figure}

For dynamic threshold, the thresholds for each category are initialized with the same value $\delta^0$ at first, which will be updated every iteration by:
\begin{equation}
\delta^{t}_c \leftarrow \beta_t \cdot \delta^{t-1}_c + (1-\beta_t) \cdot \epsilon \cdot (\overline{l}^t_c)^{\frac{1}{2}},
\label{threshupdate}
\end{equation}
where $\delta^t_c$ denotes the threshold of category $c$ at time step t, $\overline{l}^t_c$ is the mean predicted scores of the category $c$ at time step $t$, $\beta_t$ represents the update rate of dynamic threshold, and  $\epsilon$ provides a linear projection. 
Furthermore, the threshold $\delta^t_c$ will not change if class $c$ does not exist in prediction $\hat{y}^t$, and we set a fixed upper and lower bound 
$\delta_{max}$ and $\delta_{mini}$.
For unlabeled data from dynamic environments, this mechanism effectively generates an appropriate threshold for each category. Furthermore, Figure \ref{fig:dt} visualizes this process for the "car" category to provide intuition. 
As the environment shifts from ``motion'' to ``snow'', the model's average confidence (blue line) drops significantly. 
The dynamic threshold (red line) adaptively follows this trend, lowering itself to continue accepting high-quality pseudo-labels relative to the new, more challenging domain. 
Conversely, when the environment changes to ``Brightness'', the model's confidence increases, and the threshold rises accordingly.

\subsubsection{\textbf{Adaptive Randomized Restoration}}

Long-term adaptation to dynamic scenarios can lead to catastrophic forgetting, where self-training may reinforce erroneous predictions.
Even worse, the model might fail to recover when transitioning to a new domain.
Existing methods address catastrophic forgetting through stochastic reset 
\citep{wang2022continual} or pure data-driven restoration \citep{brahma2023probabilistic}. 
Nevertheless, the former may reset valuable parameters, potentially erasing essential knowledge relevant to the current domain.
On the other hand, the latter restores the least important parameters, which may retain noise parameters, thereby leading to error accumulation.
Moreover, PETAL \citep{brahma2023probabilistic}, designed for classification tasks, resets the least active parameters globally across the entire network. 
This global strategy can overlook the non-uniform scales of weights in different layers, particularly in deeper neural networks.
Our approach refines this for object detection by identifying inactive parameters within each layer individually. 
This layer-specific technique accounts for varying parameter scales at different model depths.
Building upon this, our proposed ARR mechanism further incorporates randomness to reset irrelevant parameters with a higher probability, thereby enhancing stability and robustness during restoration.

Let $p_{\theta}(y|x)$ denote the distribution over model prediction $y$, parameterized by $\theta \in \mathbb{R}^{|\theta|}$, given an input $x$.
The significance of a parameter can be determined by measuring the impact its perturbation has on the model's output.
The KL divergence $\mathcal{L}_{kl}(p_{\theta}(y|x) \parallel p_{\theta+\delta}(y|x))$ can be used to measure the sensitivity of this distribution to a small parameter perturbation $\delta \in \mathbb{R}^{|\theta|}$.  
\citet{pascanu2013revisiting,martens2020new} shows that as $\delta \rightarrow 0$, the following second-order approximation holds:
\begin{equation}
\mathbb{E}_{x}\left [ \mathcal{L}_{kl}(p_{\theta}(y|x) \parallel p_{\theta+\delta}(y|x)) \right] = \delta^{T} F_{\theta} \delta + O(\delta^{3}),
\end{equation}
where $F_{\theta} \in \mathbb{R}^{|\theta|\times|\theta|}$ is the Fisher information matrix (FIM) \citep{fisher1922mathematical}, defined as:
\begin{equation}
F_{\theta} = \mathbb{E}_{x} \left [ \mathbb{E}_{y \sim p_{\theta}(y|x)} \nabla_{\theta} \text{log}p_{\theta}(y|x) \nabla_{\theta} \text{log}p_{\theta}(y|x)^{T} \right].
\label{eqF}
\end{equation}
This approximation demonstrates that the FIM links parameter perturbations $\delta$ to the resultant changes in the model's output distribution, indicating their importance.
Consequently, we leverage FIM to guide the parameter restoration process.
However, the dimensionality of FIM $|\theta| \times |\theta|$ makes it intractable to compute in practice.
Therefore, consistent with prior work \citep{kirkpatrick2017overcoming}, we adopt the diagonal approximation of FIM, represented as a vector in $\mathbb{R}^{|\theta|}$. 
Moreover, in practice, machine learning models are typically trained on a finite set of N training samples $\{(x_j,y_j)\}_{j=1}^N$, rather than having direct access to the true data distribution $p(x)$. 
In such case, the diagonal FIM can be empirically approximated:
\begin{equation}
\hat{F_{\theta}} = \frac{1}{N} \sum_{j=1}^{N} (\nabla_{\theta} \text{log}p_{\theta}(y_j|x_j))^2,
\label{eqFhat}
\end{equation}
where $\hat{F_{\theta}} \in \mathbb{R}^{|\theta|}$ is the empirical diagonal FIM.
Each component of $\hat{F_{\theta}}$ corresponds to an individual parameter, and a larger value for such a component signifies greater influence of the corresponding parameter on the model's predictions.

For CTTA, given a batch of $N$ pairs of unlabeled test inputs and corresponding teacher's pseudo-label $\{(x_j^t,\hat{y}_j^t)\}_{j=1}^N$ at time step $t$, the FIM for the student model is defined as:
\begin{equation}
\hat{F}_{{{\theta}}_S^{t-1}} = \frac{1}{N} \sum_{j=1}^{N} (\nabla_{{{\theta}}_S^{t-1}} \text{log}p_{{{\theta}}_S^{t-1}}(\hat{y}^t_j|x^t_j))^2.
\label{fim}
\end{equation}
Since the FIM is derived from pseudo-labels, which can be noisy in the target domain, the FIM itself might assign high importance to parameters associated with these erroneous predictions.
Therefore, given a specific layer with parameters (still denoted as ${\hat{\theta}}_S^{t-1}$ for simplicity), ARR integrates a stochastic competent by employing a random matrix $R^t \sim\operatorname{Uniform}(0,1)$ with the same shape as the parameters, where each element follows a uniform distribution between 0 and 1.
The reset scores $W^t$ are then obtained by an element-wise multiplication of the FIM and the random matrix.
The parameters are updated through:
\begin{eqnarray}
    \label{score}
    &&W^t = \hat{F}_{{\theta}_S^{t-1}} \odot R^t,\\
    \label{mask}
    &&M^t=W^t < \eta, \\
    \label{weight}
    &&{\theta}_S^{t}=M^t \odot {\theta}^0 + (1-M^t) \odot {\hat{\theta}}_S^{t-1},
\end{eqnarray}
where ${\hat{\theta}}_S^{t-1}$ denotes the student model parameters after gradient descent from ${\theta}_S^{t-1}$ at time step $t$, $\odot$ is the element-wise multiplication, $M^t$ indicates the mask matrix, $<$ represents element-wise less than operation, and $\eta$ is the threshold value acquired by the $q$-quantile of $W^t$: $\eta = quantile(W^t, q)$. Consequently, the elements in $M^t$ are set to 1, when the corresponding score value is less than $\eta$, indicating the associated parameters should be reset to the source parameters ${\theta}^0$.
This strategy enables the model to retain essential knowledge while introducing randomness to enhance robustness to mitigate forgetting.

\section{EXPERIMENTAL SETUPS}
\label{s4}

In this study, we rigorously evaluate our methodology across four benchmark tasks in CTTA object detection. 
These tasks encompass continual adaptation to synthetic and real-world distribution shifts, evaluated over short and long-term periods. 
Inspired by the foundational work CoTTA \citep{wang2022continual}, the short-term CTTA tasks entails the sequential adaptation to various target domains once.
In contrast, the long-term CTTA tasks involves continually adapting the model toward a group of target domains cyclically.
\subsection{Datasets}
\paragraph{\textbf{Cityscapes}} The Cityscapes \citep{cordts2016cityscapes} is collected for urban scene understanding, encompassing 2,975 training images and 500 validation images with eight categories, i.e. person, rider, car, truck, bus,
train, motorcycle, and bicycle. 
We utilize the model pre-trained on this training set as the source model, and the data is discarded during adaptation.

\paragraph{\textbf{Cityscapes-C}}
\citet{hendrycks2019benchmarking} initially design to assess robustness against various corruptions, introducing 15 types of corruption with 5 severity levels.
We create Cityscapes-C by applying these corruptions at the maximum severity level to the validation set of clean cityscapes, treating each corruption as an individual target domain comprising 500 images.
Our short-term CTTA tasks selectively focuses on the latter 12 corruptions, including Defocus Blur, Frosted Glass Blur, Motion Blur, Zoom Blur, Snow, Frost, Fog, Brightness, Contrast, Elastic, Pixelate, and JPEG.
For the long-term CTTA task, we prioritize the five corruptions related to autonomous driving scenarios as the target domain group following \citep{gan2023cloud}, namely Fog, Motion, Snow, Brightness, and Defocus.
We repeat adaptation to the target domain group 10 times to evaluate long-term performance.

\paragraph{\textbf{SHIFT}} The SHIFT \citep{sun2022shift} is a synthetic dataset for autonomous driving, featuring real-world environmental changes. SHIFT can be categorized as clear, cloudy, overcast, rainy, and foggy, where each condition contains images taken at various times ranging from daytime to night.
For SHIFT, short-term CTTA tasks are considered.
We designate the clear condition as the source domain, with the remaining four conditions as the target domain groups including nearly 20k images in total. 

\paragraph{\textbf{ACDC}} The ACDC \citep{sakaridis2021acdc} shares the same class types as Cityscapes and is collected in four different adverse visual conditions, including Fog, Night, Rain, and Snow.
Following \citep{wang2022continual}, we use these four conditions as the target domain group for the long-term CTTA task, with 400 unlabeled images per condition. 
Similarly, the source model is continually adapted to the target domain group for 10 cycles.

\subsection{Implementation Details}
We adopt the Faster R-CNN with ResNet50 \citep{he2016deep} pre-trained on ImageNet \citep{krizhevsky2012imagenet} as the backbone.
Following \citep{vs2023towards}, we maintain a batch size of 1 to emulate a real-world application scenario where the detector adapts toward a continuous influx of images.
The source models are trained using an SGD optimizer with a learning rate of 0.001 and a momentum of 0.9.
Algorithms are implemented leveraging the Detectron2 \citep{wu2019detectron2}. 
The metric of mAP at an IoU threshold of 0.5 (mAP\@0.5) is employed for evaluation.
Each experiment is conducted on 1 NVIDIA A800 GPU.

\subsection{Baselines and Compared Approaches}
We compare AMROD with seven source-free baselines across various settings, including Source \citep{ren2015faster}, Tent \citep{wang2020tent}, IRG \citep{vs2023instance}, MemCLR \citep{vs2023towards},  CoTTA \citep{wang2022continual}, SVDP \citep{yang2024exploring}, {WHW \citep{yoo2024and},} and AMROD-upstop.
Specifically, ``Source'' represents the source model without adaptation.
Tent \citep{wang2020tent} updates the affine parameters through entropy minimization in TTA. Furthermore,
Memclr and IRG are SFDA object detection methods, also referred to as TTA methods \citep{wang2022continual}. 
MemCLR integrates cross-attention with CL, while IRG incorporates instance relation graph and CL.
CoTTA utilizes a fixed threshold for pseudo-labeling and random neuron recovery to tackle CTTA, SVDP explores sparse visual prompts for CTTA dense prediction, {and WHW utilizes source feature statistics to determine whether to update the lightweight adapter integrated into
the backbone to address CTTA}.
``AMROD-unstop'' represents AMROD without dynamic skipping.

\section{Results and Analysis}
\label{s5}
\subsection{Synthetic Continual Distribution Shift}
\subsubsection{Short-term CTTA Tasks Results}
We first evaluate the effectiveness of AMROD on the short-term Cityscapes-to-Cityscapes-C adaptation tasks. 
As depicted in Table \ref{c1}, 
Tent \citep{wang2020tent} undergoes a slight decline in performance, dropping from 15.1 to 14.2 mAP relative to the source model. This downturn may be attributed to its dependency on a large batch size to update the parameters of the batchnorm layer, making it suboptimal for online adaptation. In contrast,
CoTTA, IRG, and Memclr exhibit enhancements in performance, achieving 16.0, 17.5, and 18.0 mAP respectively.
Although the SFDA methods like IRG and MemCLR assume a static target domain, employing a large momentum update rate $\alpha$ for the teacher enables a relatively stable adaptation in the short term. 
Consequently, CL leads to a better performance than CoTTA which solely employs weight-averaged predictions.
Moreover, SVDP, which utilizes visual prompts, achieves a sub-optimal performance of 19.1 mAP.
{Furthermore, while WHW selectively updates lightweight adapters using fewer adaptation iterations, this strategy might skip necessary adaptation steps, thereby limiting its performance to 20.0 mAP.}
Remarkably, our proposed method improves the performance to 20.8 mAP, consistently outperforming the above approaches. 
Additionally, the dynamic skipping strategy in AM reduces adaptation iterations by 20\% while achieving a 0.2 mAP improvement compared to AMROD-unstop.
AMROD ensures a more reliable and efficient adaptation to target domains characterized by intense changes in the short term.   

\begin{table}[tbp]
\setlength\tabcolsep{3.0pt} 
\centering
\caption{Experimental results (mAP\@0.5) and adaptation iterations of Cityscapes-to-Cityscapes-C short-term CTTA task. We evaluate the performance by continually adapting the source model to twelve corruptions. ``-unstop'' denotes AMROD without skipping.
}
\resizebox{\textwidth}{!}{
\begin{tabular}{l|cccccccccccc|ccc}
\toprule
Time          & \multicolumn{12}{c|}{$t\xrightarrow{\hspace{0.85\linewidth}}$}                                                                                                                                                                     & \multicolumn{3}{c}{All}      \\ \midrule
Condition     & Defocus      & Glass         & Motion        & Zoom         & Snow         & Frost         & Fog           & Brightness    & Contrast      & Elastic       & Pixelate      & Jpeg          & Mean          & Gain  &Iter.       \\ \midrule
Source         & 6.8          & 8.1           & 8.0             & 1.5          & 0.2          & 6.8           & 34.6          & 30.7          & 3.0             & 50.2          & 17.6          & 13.5          & 15.1        & /   & /         \\
Tent         & 6.8          & 7.8           & 7.7           & 1.3          & 0.2          & 6.1           & 33.1          & 28.0            & 2.2           & 51.1 & 14.8          & 11.0            & 14.2          & -0.9   & 6.0k
\\
CoTTA         & 7.8          & 9.0             & 8.9           & 1.8          & 0.3          & 7.1           & 38.4          & 31.1          & 8.6           & 49.6          & 16.2          & 13.1          & 16.0            & +0.9   & 6.0k      \\
SVDP         & 7.7          & 10.1             & 9.7           & 2.3          & 0.7          & 13.0           & 42.4          & 45.2          & 15.4           & 47.2          & 21.2          & 14.8          & 19.1            & +4.0       & 6.0k  \\
IRG           & 8.0            & 11.0            & 9.3           & 3.4          & 1.2          & 13.0            & 37.9          & 41.3          & 15.9          & 38.9          & 16.9          & 13.4          & 17.5          & +2.4      & 6.0k   \\
MemCLR        & 8.5          & 10.4          & 10.6          & 2.7          & 1.1          & 12.2          & 41.4          & 41.6          & 16.4          & 43.1          & 15.4          & 12.7          & 18.0            & +2.9       & 6.0k  \\ 
 WHW & \textbf{9.2} &9.3 &12.2 &1.9 &0.8 & 14.9 & 44.6 & \textbf{48.9} & 10.5 & \textbf{51.9} & 20.0 & 15.3 & 20.0 & +4.9 & 4.1k  \\\midrule
Ours-unstop & 8.6 & \textbf{12.1} & 11.7 & \textbf{3.6} & \textbf{1.5} & \textbf{16.7} & \textbf{44.7} & 48.1 & 16.7 & 47.4          & 22.5 & 13.9 & 20.6 & +5.5  & 6.0k \\ 
\rowcolor{lightgreen}
\textbf{Ours} & 8.7 & 11.7 & \textbf{12.4} & \textbf{3.6} & \textbf{1.5} & 13.5 & 43.2 & 47.3 & \textbf{18.4} & 47.0          & \textbf{24.8} & \textbf{17.2} & \textbf{20.8} & \textbf{+5.7} & \textbf{4.7k} \\
\bottomrule

\end{tabular}
}

\label{c1}
\centering

\end{table}

\begin{table*}[tbp]
\centering
\caption{Experimental results (mAP\@0.5) and adaptation iterations of Cityscapes-to-Cityscapes-C long-term CTTA task. 
We evaluate the performance by continually adapting the source model to the five corruption ten times. 
``-unstop'' denotes AMROD without skipping.
To save space, we display selected rounds of results.
}
\resizebox{\textwidth}{!}{
\setlength\tabcolsep{1.8pt} 
\begin{tabular}{l|ccccccccccccccc|ccc}
\toprule
Time      & \multicolumn{15}{c|}{$t\xrightarrow{\hspace{0.98\linewidth}}$}                                                                                                                                                                                                                                                & \multicolumn{2}{c}{}           \\ \midrule
Round     & \multicolumn{5}{c|}{1}                                                                          & \multicolumn{5}{c|}{5}                                                                          & \multicolumn{5}{c|}{10}                                            & \multicolumn{3}{c}{All}        \\ \midrule
Condition & Fog         & Motion        & Snow         & Brightness    & \multicolumn{1}{c|}{Defocus}       & Fog         & Motion        & Snow         & Brightness    & \multicolumn{1}{c|}{Defocus}       & Fog  & Motion      & Snow          & Brightness    & Defocus       & Mean          & Gain  &Iter.         \\ \midrule
Source    & 36.1          & 8.1           & 0.2          & 31.0          & \multicolumn{1}{c|}{6.7}           & 36.1          & 8.1           & 0.2          & 31.0            & \multicolumn{1}{c|}{6.7}           & 36.1          & 8.1         & 0.2           & 31.0            & 6.7           & 16.4          & / & /             \\
Tent      & 35.8          & 8.1           & 0.2          & 29.2        & \multicolumn{1}{c|}{6.2}           & 30.7          & 10.2          & 0.8          & 27.9          & \multicolumn{1}{c|}{12.4}          & 20.4          & 5.0           & 0.1           & 11.5          & 2.4           & 12.0            & -4.4    &25.0k       \\
CoTTA     & 38.2          & \textbf{10.6} & 0.4          & 33.8        & \multicolumn{1}{c|}{9.3}           & 40.8          & 10.4          & 0.6          & 36.5          & \multicolumn{1}{c|}{9.2}           & 40.8          & 10.4        & 0.5           & 36.3          & 9.6           & 19.1          & +2.7  &25.0k         \\
SVDP     & 36.8          & 8.8 & 0.6          & 43.5        & \multicolumn{1}{c|}{11.2}           & 45.0          & 15.4          & 3.8          & 47.8          & \multicolumn{1}{c|}{19.9}           & 41.5          & 16.2        & 4.6           & 44.6          & 20.2           & 25.3          & +8.9        &25.0k   \\
IRG       & 37.9          & 9.0             & 0.7          & 44.3        & \multicolumn{1}{c|}{12.2}          & 45.6          & 17.7          & 6.0            & 46.7          & \multicolumn{1}{c|}{21.9}          & 37.0            & 16.7        & 7.1           & 38.5          & 20.5          & 25.6          & +9.2          &25.0k \\
MemCLR    & 37.7          & 8.9           & \textbf{0.8} & 45.5        & \multicolumn{1}{c|}{13.5}          & 45.0            & 18.1          & 5.1          & 46.8          & \multicolumn{1}{c|}{22.4}          & 36.8          & 18.5        & 7.5           & 38.4          & 21.8          & 26.0            & +9.6         &25.0k  \\ 
WHW &\textbf{40.8} &9.9 &0.7 &43.4 &\multicolumn{1}{c|}{12.5} &42.8 &15.5 &2.0 &46.9 &\multicolumn{1}{c|}{17.9} &36.7 &18.3 &3.5 &44.1 &16.2&24.2 &+7.8 &18.0k\\ \midrule
Ours-unstop      & 39.0 & 10.4          & \textbf{0.8} & \textbf{48.0} & \multicolumn{1}{c|}{\textbf{13.8}} & \textbf{49.4} & \textbf{18.6} & 7.7 & 51.7 & \multicolumn{1}{c|}{\textbf{25.0}} & 45.7 & \textbf{20.0} & \textbf{11.8} & 46.4 & 25.8 & 29.0 & +12.6 &25.0k \\ 
\rowcolor{lightgreen}
\textbf{Ours}      & 39.2 & 9.9          & \textbf{0.8} & 46.7 & \multicolumn{1}{c|}{12.7} & \textbf{49.4} & 18.5 & \textbf{8.0} & \textbf{52.9} & \multicolumn{1}{c|}{24.7} & \textbf{46.2} & 19.5 & 11.3 & \textbf{47.3} & \textbf{29.1} & \textbf{29.2} & \textbf{+12.8} &\textbf{20.1k} \\
\bottomrule
\end{tabular}
}

\label{c2}
\centering
\end{table*}

\subsubsection{Long-term CTTA Tasks Results} As presented in Table \ref{c2}, the long-term tasks outcomes reveal the source model's poor performance, with an average mAP of 16.4. 
Despite the improvement of Memclr and IRG, their performance also begins to decline in the later rounds, failing to maintain stability over long-term adaptation.
We believe this is due to the aforementioned methods not accounting for continual distribution shifts, resulting in error accumulation and catastrophic forgetting.
Furthermore, CoTTA utilizes a stochastic restoration mechanism to mitigate forgetting, but its randomness may result in losing crucial information, thus limiting its performance at 19.1 mAP.
{We also observe that WHW exhibits noticeable performance degradation over prolonged continuous shifts, dropping to an average of 24.2 mAP. 
This indicates that freezing the backbone and relying solely on the additional adapters struggles to capture evolving target knowledge over extended periods.}
In contrast, SVDP which employs sparse visual prompts raises the performance to 25.3 mAP.
Particularly, AMROD yields a remarkable 12.8 mAP enhancement over the Source and surpasses all comparative baselines using 80\% adaptation iterations, which employs the ARR mechanism to conserve valuable knowledge while eliminating noise from prior domains.
These findings empirically validate the effectiveness of our proposed method in ensuring stable and efficient adaptation amidst synthetic continual distribution shifts, across both short-term and long-term scenarios.

\subsection{Real-World Continual Distribution Shift}
\subsubsection{Short-term CTTA Tasks Results}

We also evaluate our method on the real-world continual distribution shift dataset. The results of the SHIFT short-term CTTA tasks are shown in Table \ref{t4}. 
While other baseline methods suffer from performance decline, SVDP which employs a fixed threshold for pseudo-labeling, stochastic restoration, and prompts learning achieves a 0.5 mAP improvement.
{Additionally, while WHW uses fewer adaptation iterations, it risks skipping necessary adaptations, resulting in a marginal improvement of 0.1 mAP.}
Moreover, AMROD which introduces the AM module shows a superior performance of 43.9 mAP with 16\% improvement in efficiency, {achieving a better parameter-efficiency tradeoff}.

\begin{table}[t]
\centering
\renewcommand{\arraystretch}{0.9} 
\caption{Experimental results (mAP\@0.5) of SHIFT short-term CTTA task. We evaluate the performance by continually adapting the source model to the four conditions.}
\resizebox{0.7\textwidth}{!}{
\begin{tabular}{l|cccc|ccc}
\toprule
Time          & \multicolumn{4}{c|}{$t\xrightarrow{\hspace{0.36\linewidth}}$}                                         & \multicolumn{3}{c}{All}       \\ \midrule
Condition     & Cloudy        & Overcast      & Rainy         & Foggy         & Mean          & Gain     &Iter.     \\ \midrule
Source        & 51.8          & 41.5 & 43.8          & 33.9          & 42.7          & /   & /          \\
Tent          & 50.9           & 39.5           &  36.3          & 23.1           &  37.5         & -5.2      &19.6k    \\
CoTTA         & 51.1          & 40.1          & 40.7          & 29.9          & 40.5          & -2.2     &19.6k     \\
SVDP          & 52.0            & 41.0            & 43.8          & 35.9          & 43.2          & +0.5      &19.6k    \\
IRG           & 51.9          & 40.6          & 42.7          & 34.3          & 42.4          & -0.3        &19.6k  \\
MemCLR        & 51.8          & 40.4          & 42.6          & 35.0            & 42.4          & -0.3       &19.6k   \\ 
WHW &51.7 &41.4 &43.9 &34.0 &42.8 &+0.1 &10.2k \\

\midrule
Ours-unstop & \textbf{52.3} & 41.3          & 44.1 & 36.8 & 43.6 & +0.9 &19.6k  \\ 
\rowcolor{lightgreen}
\textbf{Ours} & \textbf{52.3} & \textbf{41.6}          & \textbf{44.4} & \textbf{37.3} & \textbf{43.9} & \textbf{+1.2} &\textbf{16.4k} \\
 \bottomrule
\end{tabular}
}
\label{t4}
\centering
\end{table}

\begin{table}[tbp]
\centering
\caption{Experimental results (mAP\@0.5) and adaptation iterations of Cityscapes-to-ACDC long-term CTTA task. 
We evaluate the performance by continually adapting the source model to the four conditions ten times.
``-unstop'' denotes AMROD without skipping.
To save space, we display selected rounds of results.
}
\resizebox{\textwidth}{!}{
\setlength{\tabcolsep}{3pt}
\begin{tabular}{l|cccccccccccccccc|ccc}
\toprule
Time  & \multicolumn{16}{c|}{$t\xrightarrow{\hspace{0.9\linewidth}}$}                                                                                                                                                                                                                                                                                                  & \multicolumn{2}{c}{}          \\ \midrule
Round & \multicolumn{4}{c|}{1}                                                           & \multicolumn{4}{c|}{4}                                                             & \multicolumn{4}{c|}{7}                                                           & \multicolumn{4}{c|}{10}                                       & \multicolumn{3}{c}{All}       \\ \midrule
Condition                  & Fog           & Night         & Rain        & \multicolumn{1}{c|}{Snow}          & Fog           & Night         & Rain          & \multicolumn{1}{c|}{Snow}          & Fog           & Night         & Rain        & \multicolumn{1}{c|}{Snow}          & Fog           & Night         & Rain          & Snow          & Mean          & Gain  &Iter.        \\ \midrule
Source                      & 52.3          & 18.7          & 33.5        & \multicolumn{1}{c|}{39.6}          & 52.3          & 18.7          & 33.5          & \multicolumn{1}{c|}{39.6}          & 52.3          & 18.7          & 33.5        & \multicolumn{1}{c|}{39.6}          & 52.3          & 18.7          & 33.5          & 39.6          & 36.0            & /      & /         \\
Tent                       & 52.4          & 18.6          & 33.4        & \multicolumn{1}{c|}{38.9}          & 51.7          & 17.4          & 31.4          & \multicolumn{1}{c|}{36.0}            & 45.8          & 14.6          & 28.1        & \multicolumn{1}{c|}{28.5}          & 35.0            & 9.5           & 21.5          & 18.3          & 30.5          & -5.5      &16.0k    \\
CoTTA                      & 53.7 & 19.7          & 38.0 & \multicolumn{1}{c|}{42.4}          & 53.1          & 19.7          & 37.7          & \multicolumn{1}{c|}{42.9}          & 51.3          & 19.0            & 36.5        & \multicolumn{1}{c|}{41.8}          & 50.9          & 19.1          & 36.0            & 42.4          & 37.8          & +1.8   &16.0k       \\
SVDP         & 52.8          & 20.0           & 35.6        & \multicolumn{1}{c|}{42.0}             & 54.6 & 23.5          & 38.7 & \multicolumn{1}{c|}{43.8}          & 52.8          & 24.0          & 38.6          & \multicolumn{1}{c|}{43.9}                   &51.8           & 23.6          & 38.2            & 43.0         & 39.5          & +3.5      &16.0k   \\
IRG              & 52.7          & 20.6          & 36.0          & \multicolumn{1}{c|}{42.9} & 53.6          & 23.2          & 38.1          & \multicolumn{1}{c|}{45.6}          & 51.0            & 23.1          & 37.1        & \multicolumn{1}{c|}{42.8}          & 49.7          & 22.5          & 36.5          & 40.7          & 38.9          & +2.9        &16.0k  \\
MemCLR         & 52.9          & \textbf{21.2} & 35.2        & \multicolumn{1}{c|}{42.8}          & 52.8          & 23.2          & 38.2          & \multicolumn{1}{c|}{44.4}          & 51.4          & 23.2          & 37.1        & \multicolumn{1}{c|}{42.9}          & 49.9          & 22.9          & 36.6          & 41.0            & 38.7          & +2.7       &16.0k   \\ 
WHW &\textbf{54.5} &21.1 &\textbf{39.6} &\multicolumn{1}{c|}{\textbf{43.5}}  &\textbf{55.7} &21.3 &39.1 &\multicolumn{1}{c|}{41.5} &53.0 &21.7 &35.9 &\multicolumn{1}{c|}{39.5} &50.9 &20.9 &33.8 &38.3 &38.3 &+2.3 &8.0k\\\midrule
Ours-unstop             & 52.9          & 20.4          & 34.5        & \multicolumn{1}{c|}{42.9} & 54.5 & 23.8 & \textbf{39.6} & \multicolumn{1}{c|}{\textbf{45.8}} & \textbf{53.5} & 25.1 & \textbf{39.0} & \multicolumn{1}{c|}{45.3} &52.5 & 24.3 & \textbf{38.8} &44.7 & 40.3 & +4.3  &16.0k\\ 
\rowcolor{lightgreen}
\textbf{Ours}             & 52.8          & 20.1          & 35.2        & \multicolumn{1}{c|}{42.9} & 54.4 & \textbf{24.6} & 38.6 & \multicolumn{1}{c|}{45.5} & \textbf{53.5} & \textbf{25.6} & 38.5 & \multicolumn{1}{c|}{\textbf{46.2}} & \textbf{53.5} & \textbf{25.4} & 37.4 & \textbf{45.3} & \textbf{40.4} & \textbf{+4.4} &\textbf{13.2k}\\
\bottomrule
\end{tabular}
}
\label{acdc}
\centering
\end{table}

\subsubsection{Long-term CTTA Tasks Results}

The experimental results of the ACDC long-term CTTA tasks are summarized in Table \ref{acdc}. 
CoTTA, IRG, and MemCLR potentially suffer from error accumulation and catastrophic forgetting, manifesting in a rapid decline in performance during the later stages.
{Additionally, WHW also struggles with long-term shifts, as its performance under Fog conditions drops from 54.5 mAP in the beginning to  50.9 mAP at the end.}
Similarly, SVDP achieves a sub-optimal performance of 39.5 mAP.
Notably, AMROD achieves a performance of 40.4 mAP, yielding an absolute improvement of 0.9 mAP over baselines.
While AMROD may not display a significant advantage over other methods in the initial round, its performance continually improves and remains stable throughout the long-term adaptation.

These results underscore the capability of our method to foster robust adaptation toward the real-world continual distribution shift in both the short and long term.

    
      
    
      
    
      
    
      
  

\subsection{Qualitative Results}
\begin{figure*}[htbp]
  \centering
  \subfloat[Source]{
      \centering
      \includegraphics[trim={145 492 700 30}, clip,width=0.32\textwidth]{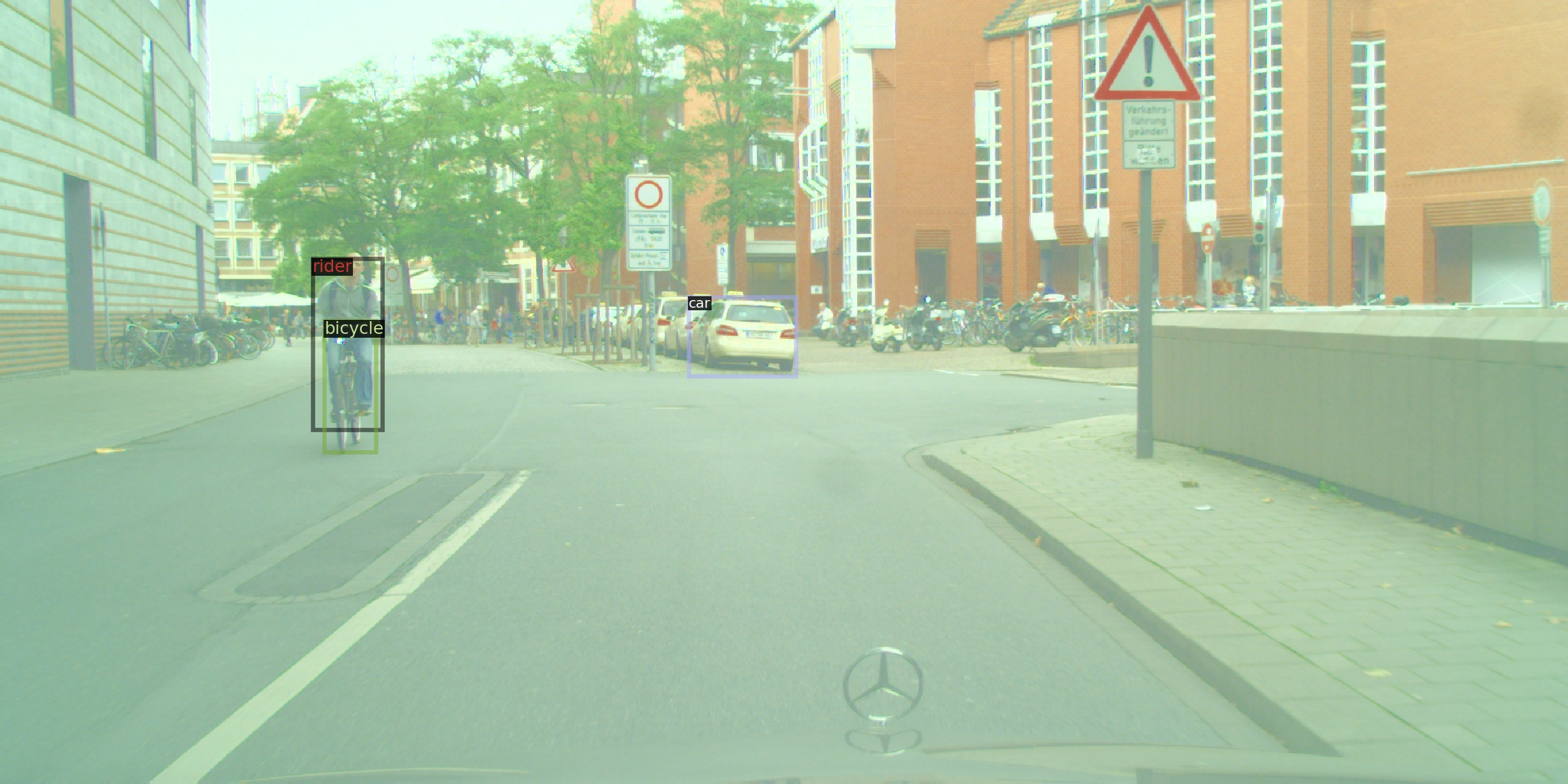}
      \label{bfig:subfig1}
  }
  \subfloat[SVDP]{
      \centering
      \includegraphics[trim={145 492 700 30}, clip,width=0.32\textwidth]{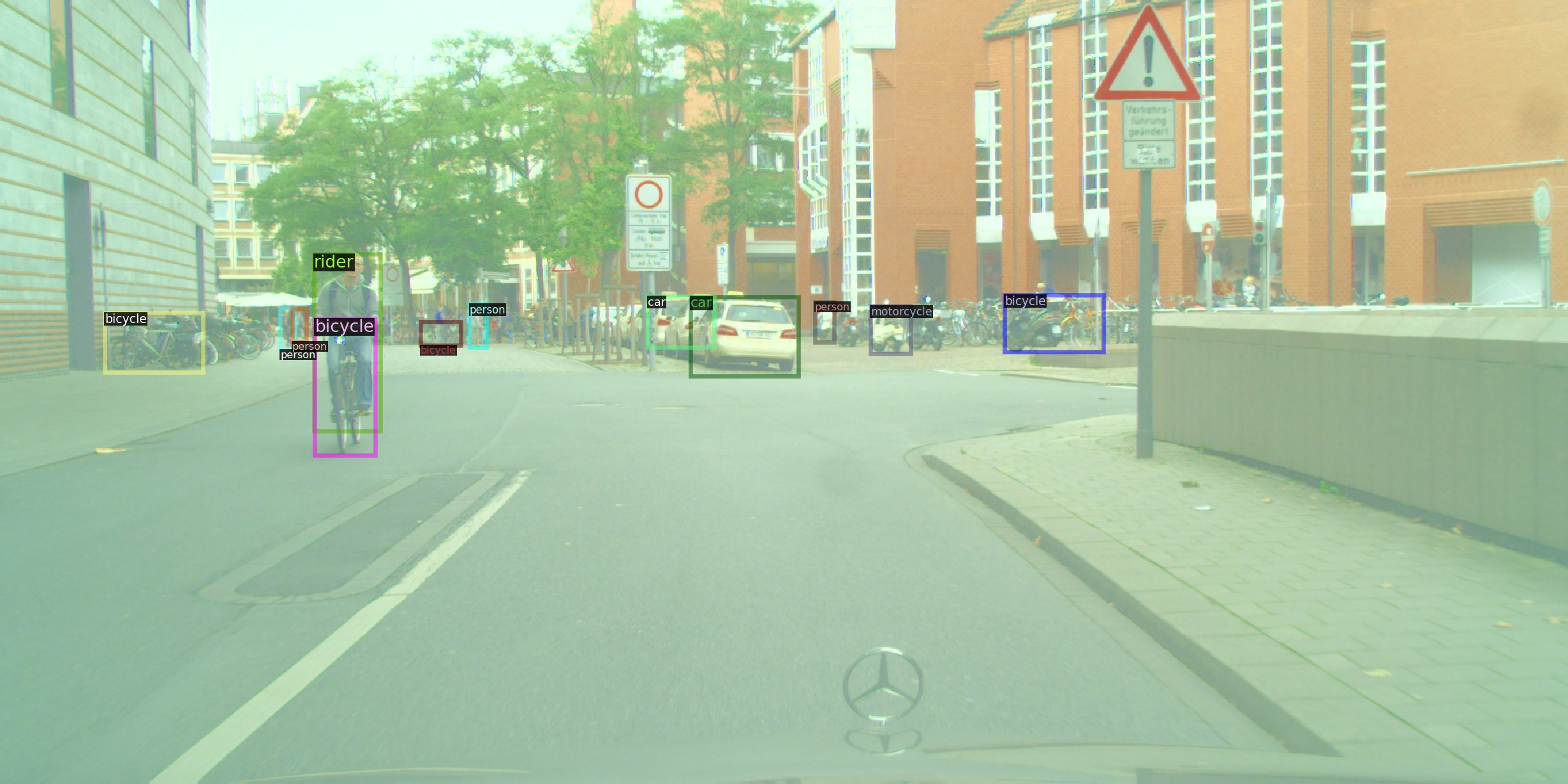}
      \label{bfig:subfig2}
  }
  \subfloat[CoTTA]{
      \centering
      \includegraphics[trim={145 492 700 30}, clip,width=0.32\textwidth]{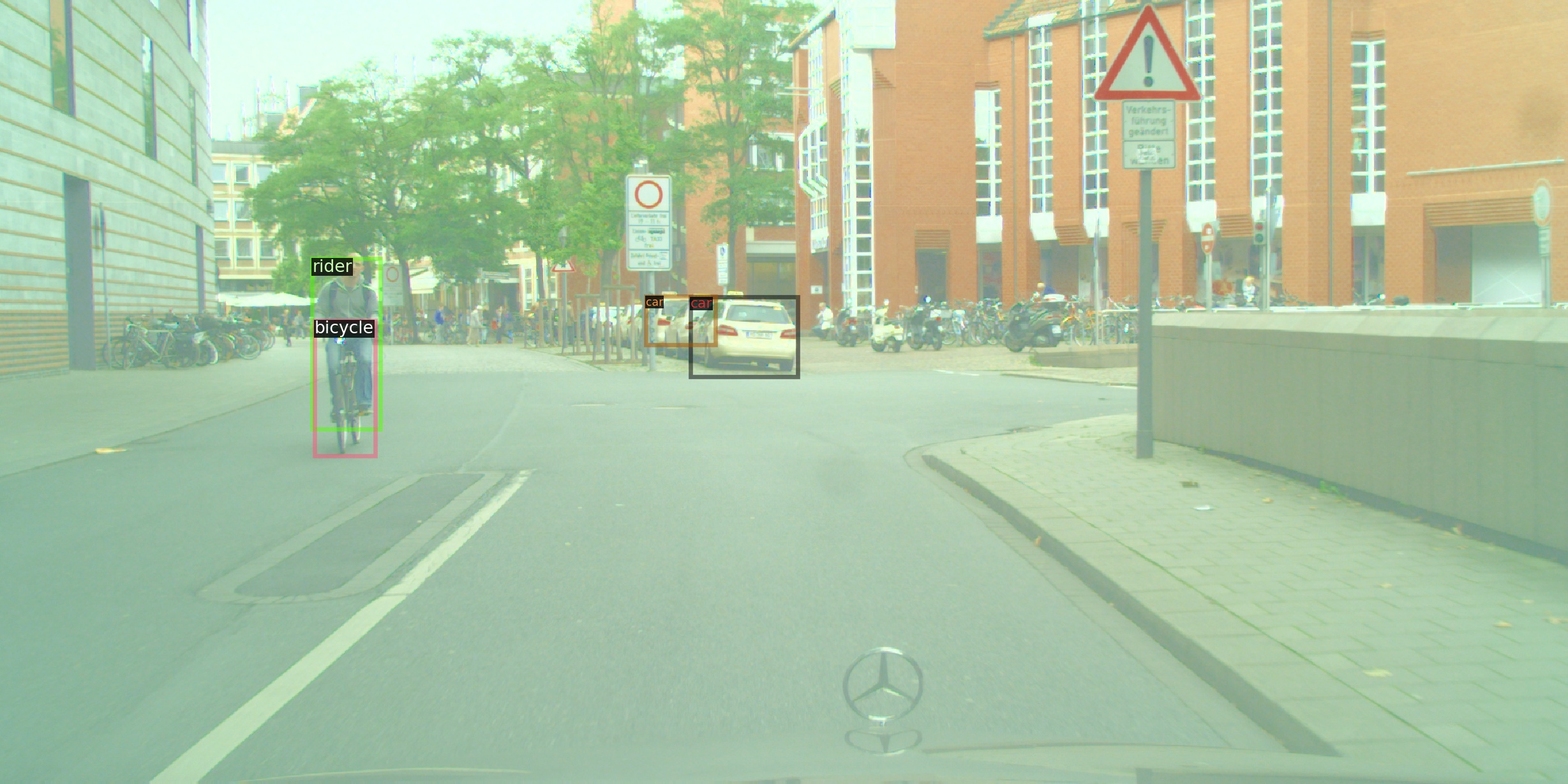}
      \label{bfig:subfig3}
  }
  
  \vspace{-8pt}
  
  \subfloat[IRG]{
      \centering
      \includegraphics[trim={145 492 700 30}, clip,width=0.32\textwidth]
      {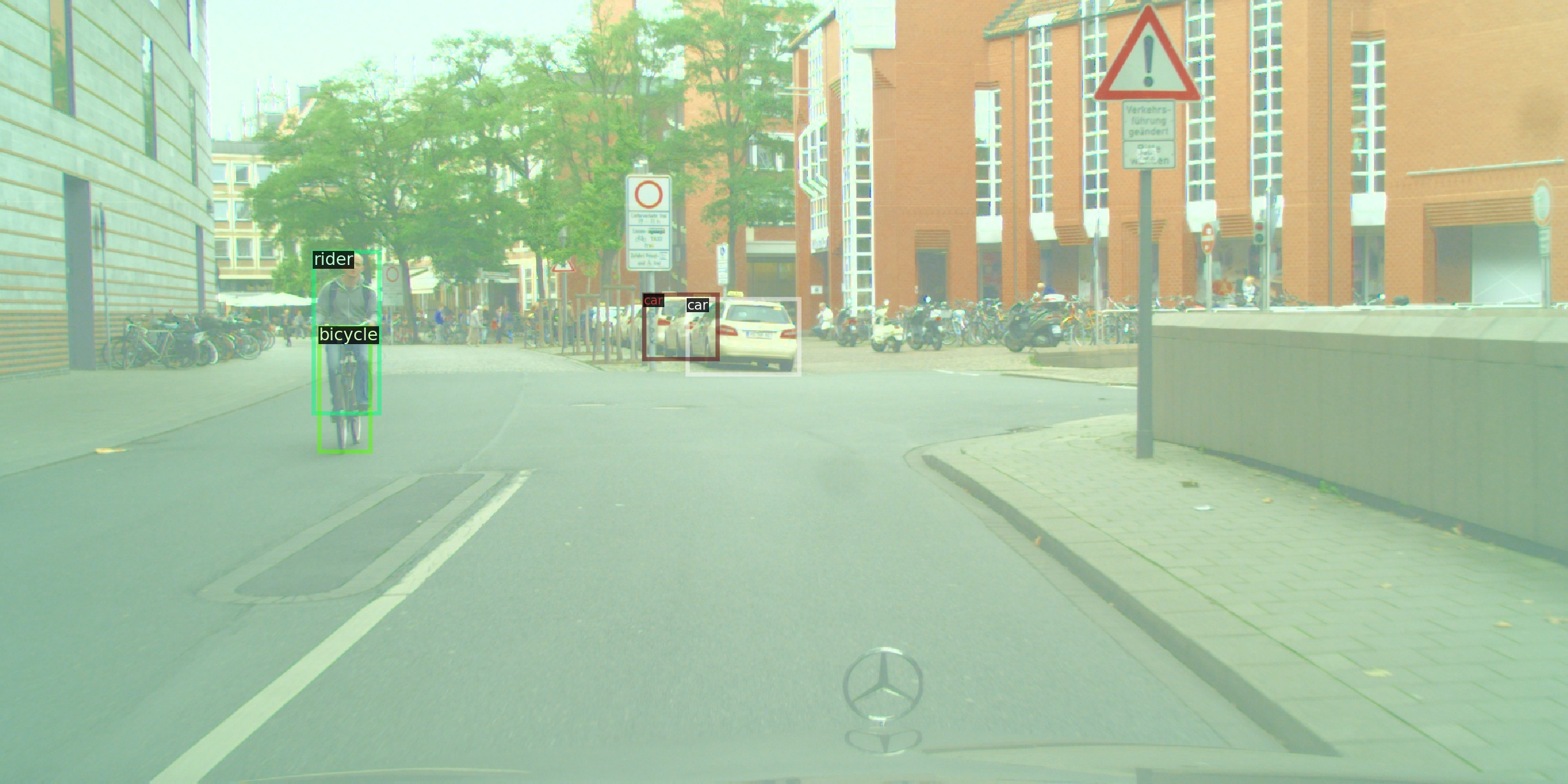}
      \label{bfig:subfig4}
  }
  \subfloat[MemCLR]{
      \centering
      \includegraphics[trim={145 492 700 30}, clip,width=0.32\textwidth]{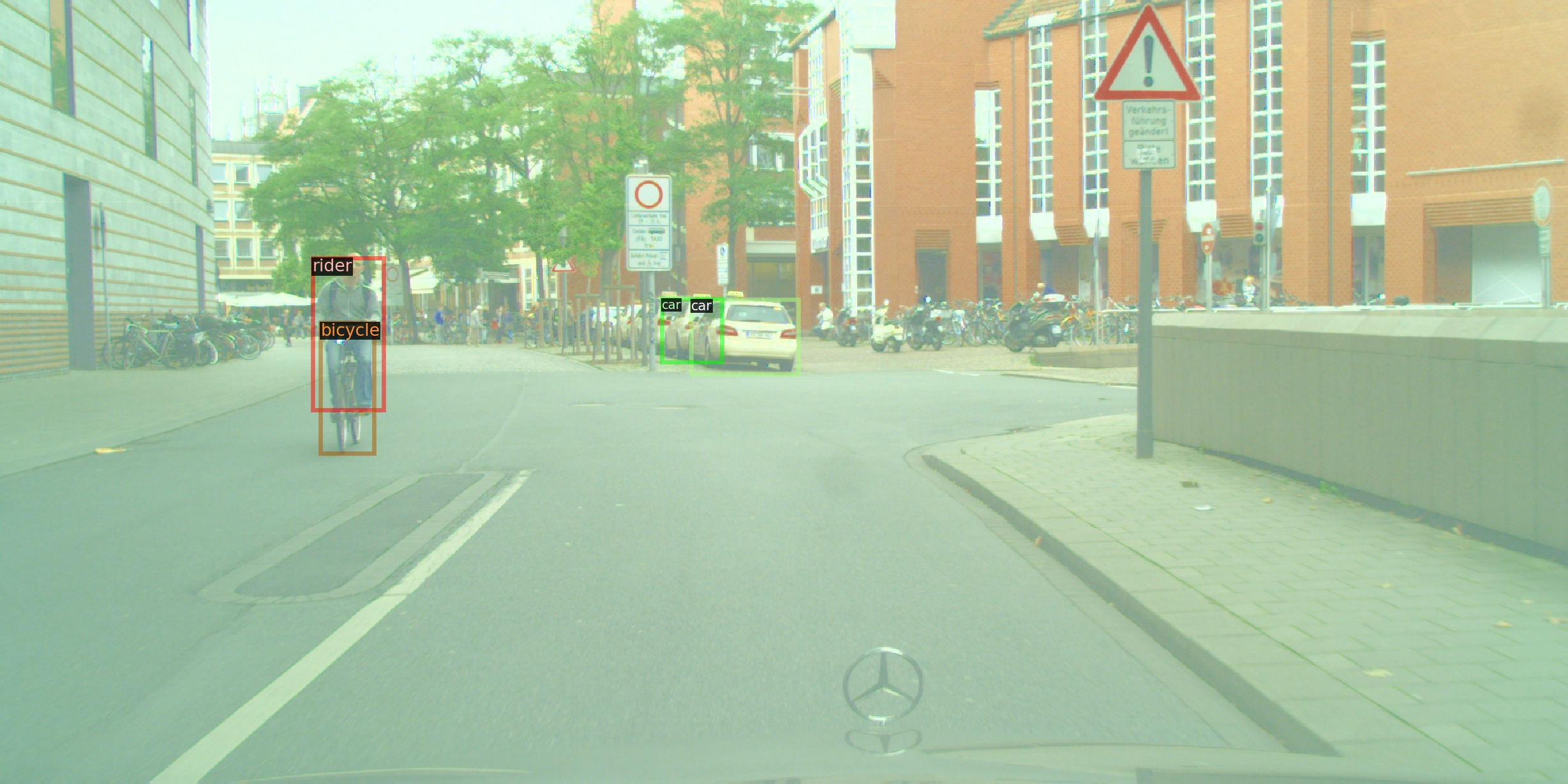}
      \label{bfig:subfig5}
  }
  \subfloat[AMROD (ours)]{
      \centering
      \includegraphics[width=0.32\textwidth]{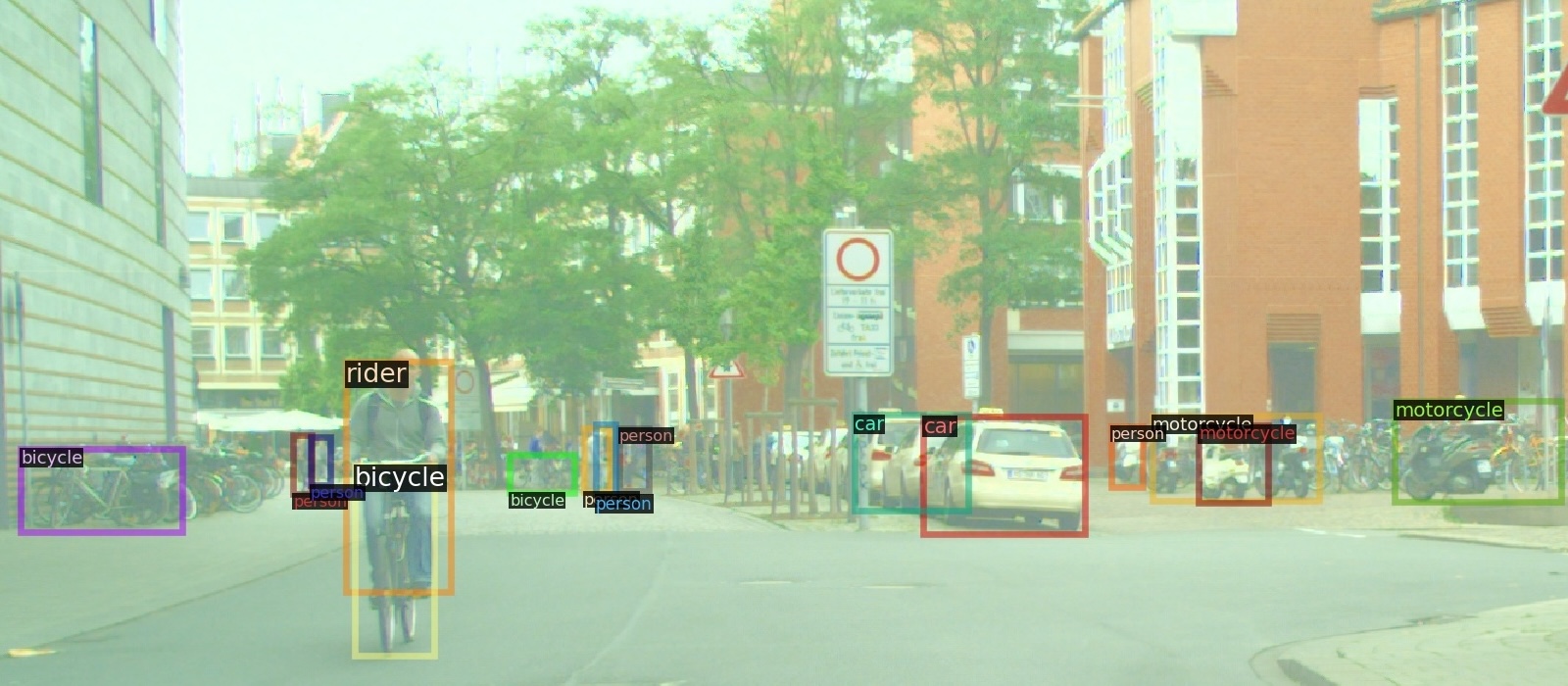}
      \label{bfig:subfig6}
  }
  \vspace{-10pt}
 \caption{Qualitative results. We compare the detection results of the AMROD and other baseline methods in the $10^{th}$ round of adaption to \textit{Brightness} corruption on the long-term Cityscapes-to-Cityscapes-C task.}
 \vspace{-10pt}
 \label{f1}
\end{figure*}

\begin{figure}[htbp]
  \centering
  \subfloat[Source]{
      \centering
      \includegraphics[trim={490 486 350 40}, clip,width=0.32\textwidth]
      {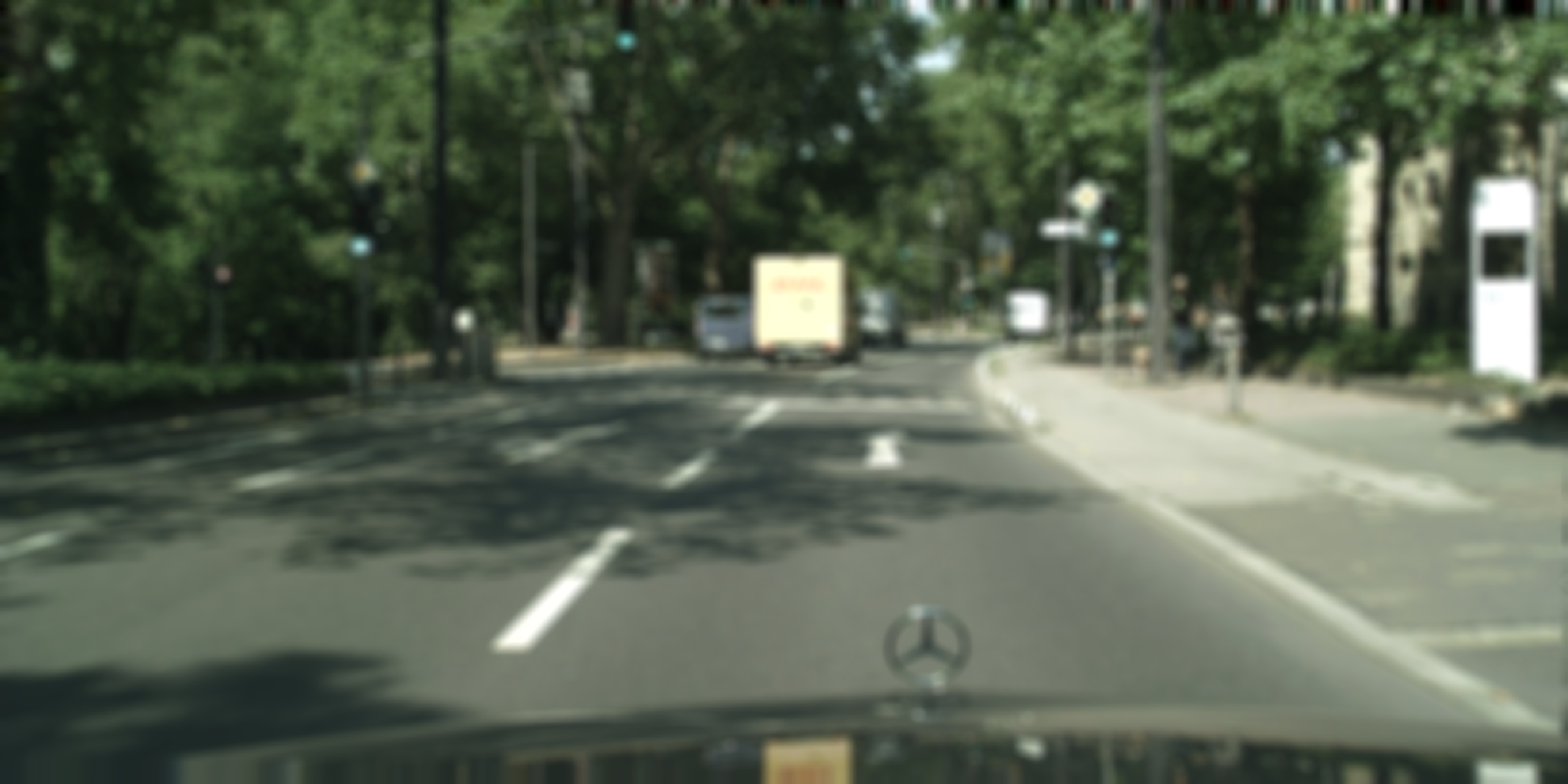}
      \label{dfig:subfig1}
  }
  \subfloat[SVDP]{
      \centering
      \includegraphics[trim={490 486 350 40}, clip,width=0.32\textwidth]{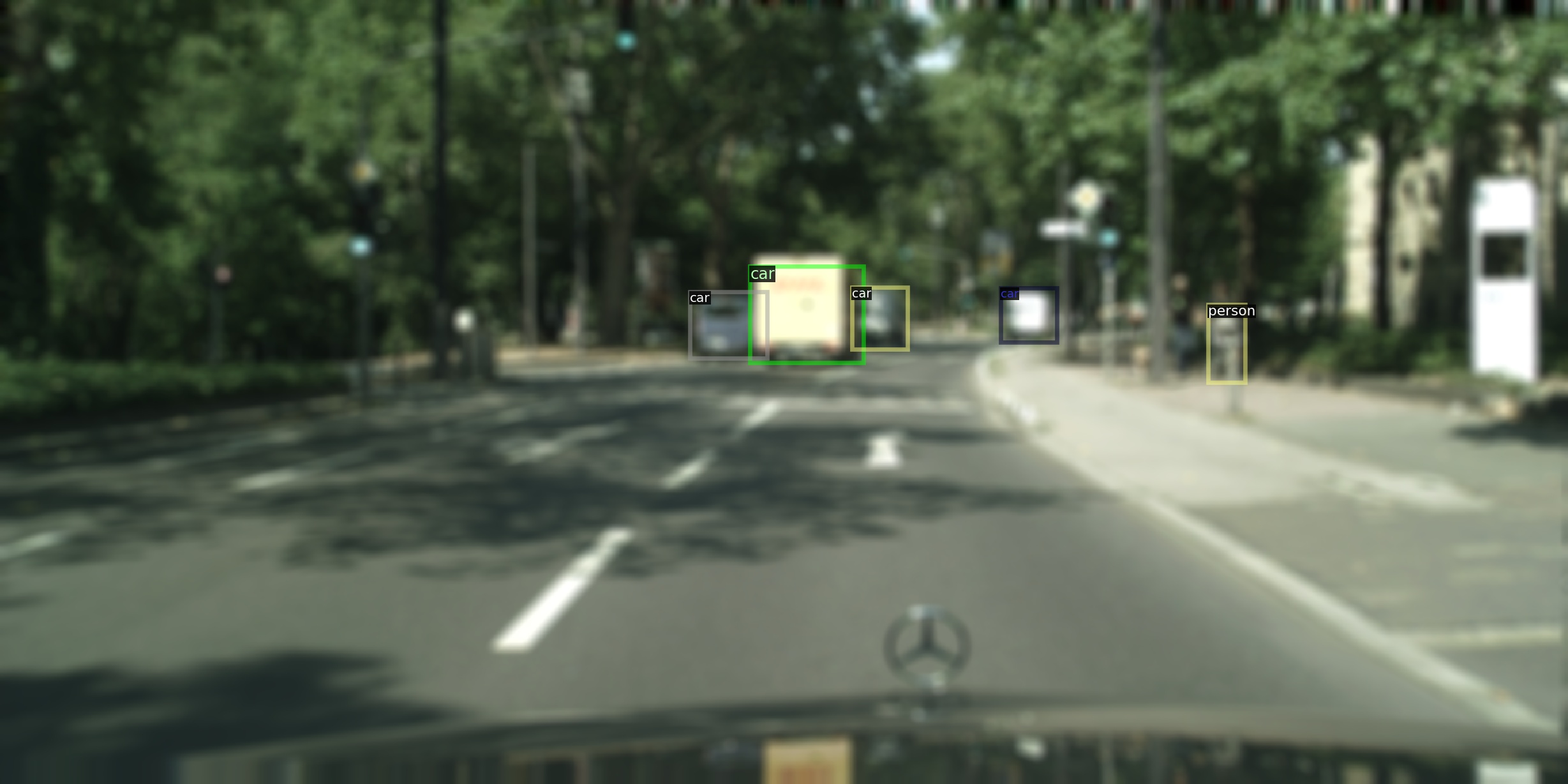}
      \label{dfig:subfig2}
  }
  \subfloat[CoTTA]{
      \centering
      \includegraphics[trim={490 486 350 40}, clip,width=0.32\textwidth]{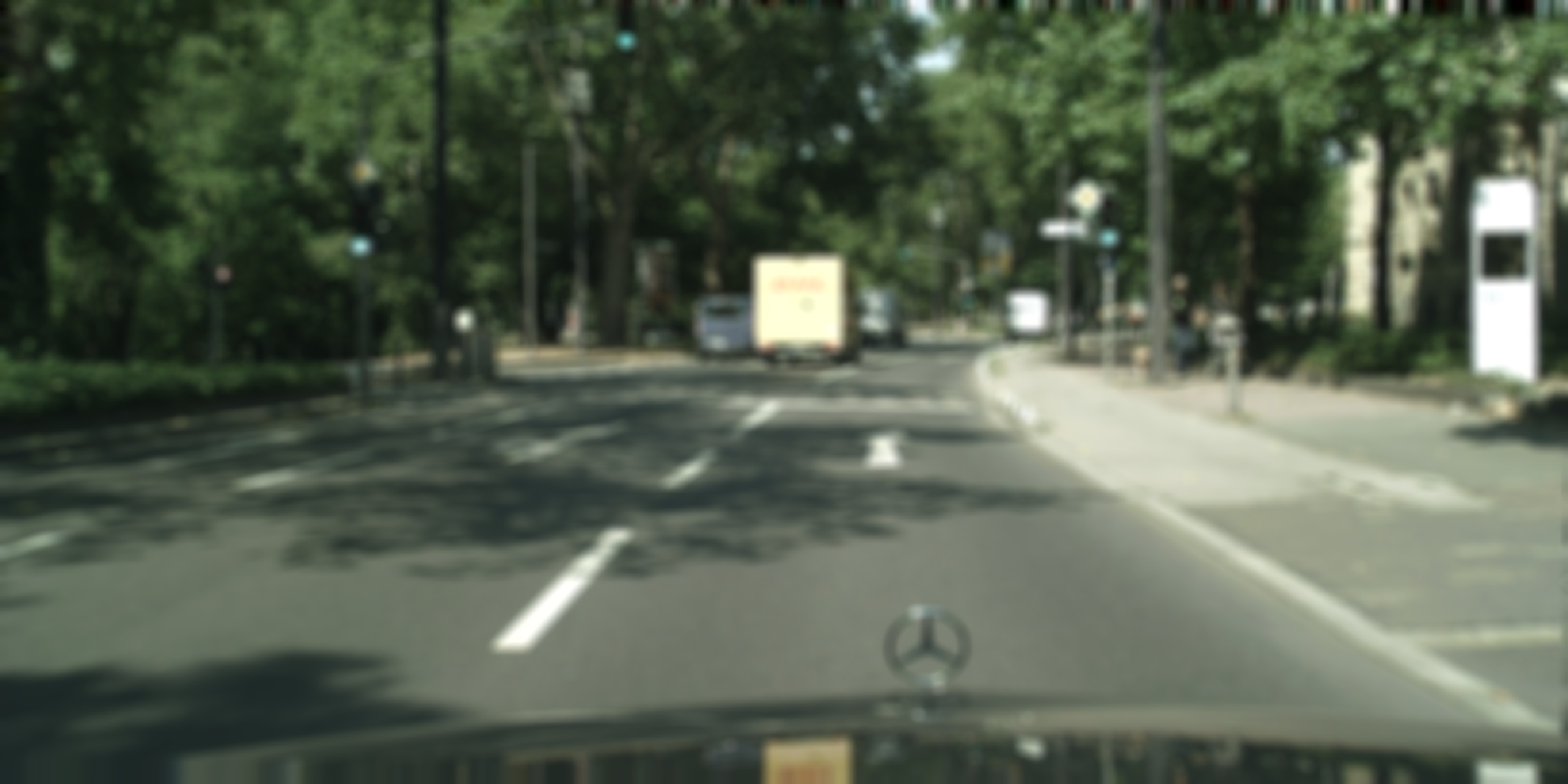}
      \label{dfig:subfig3}
  }

  \vspace{-8pt}
  
  \subfloat[IRG]{
      \centering
      \includegraphics[trim={490 486 350 40}, clip,width=0.32\textwidth]{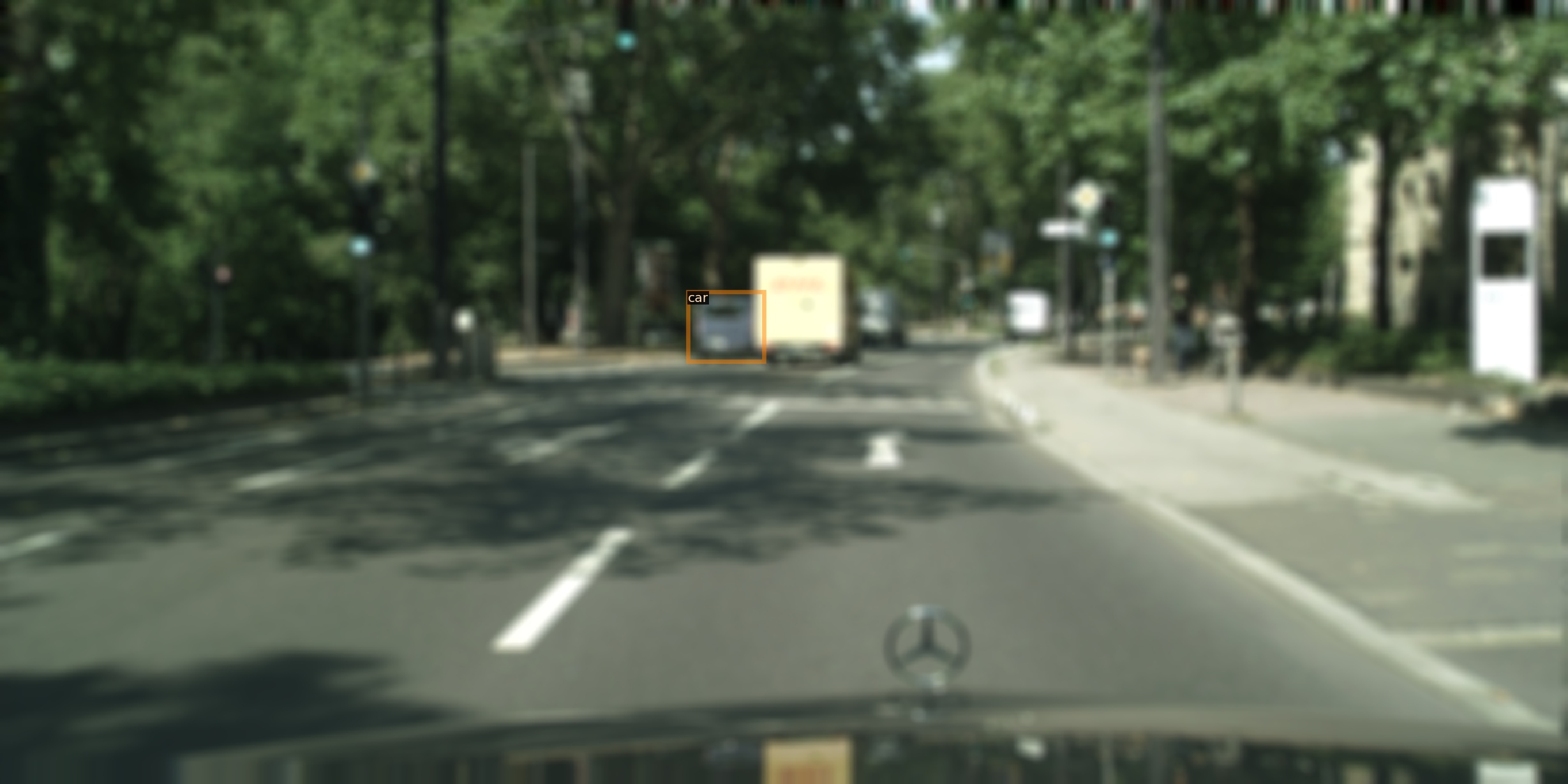}
      \label{dfig:subfig4}
  }
  \subfloat[MemCLR]{
      \centering
      \includegraphics[trim={490 486 350 40}, clip,width=0.32\textwidth]{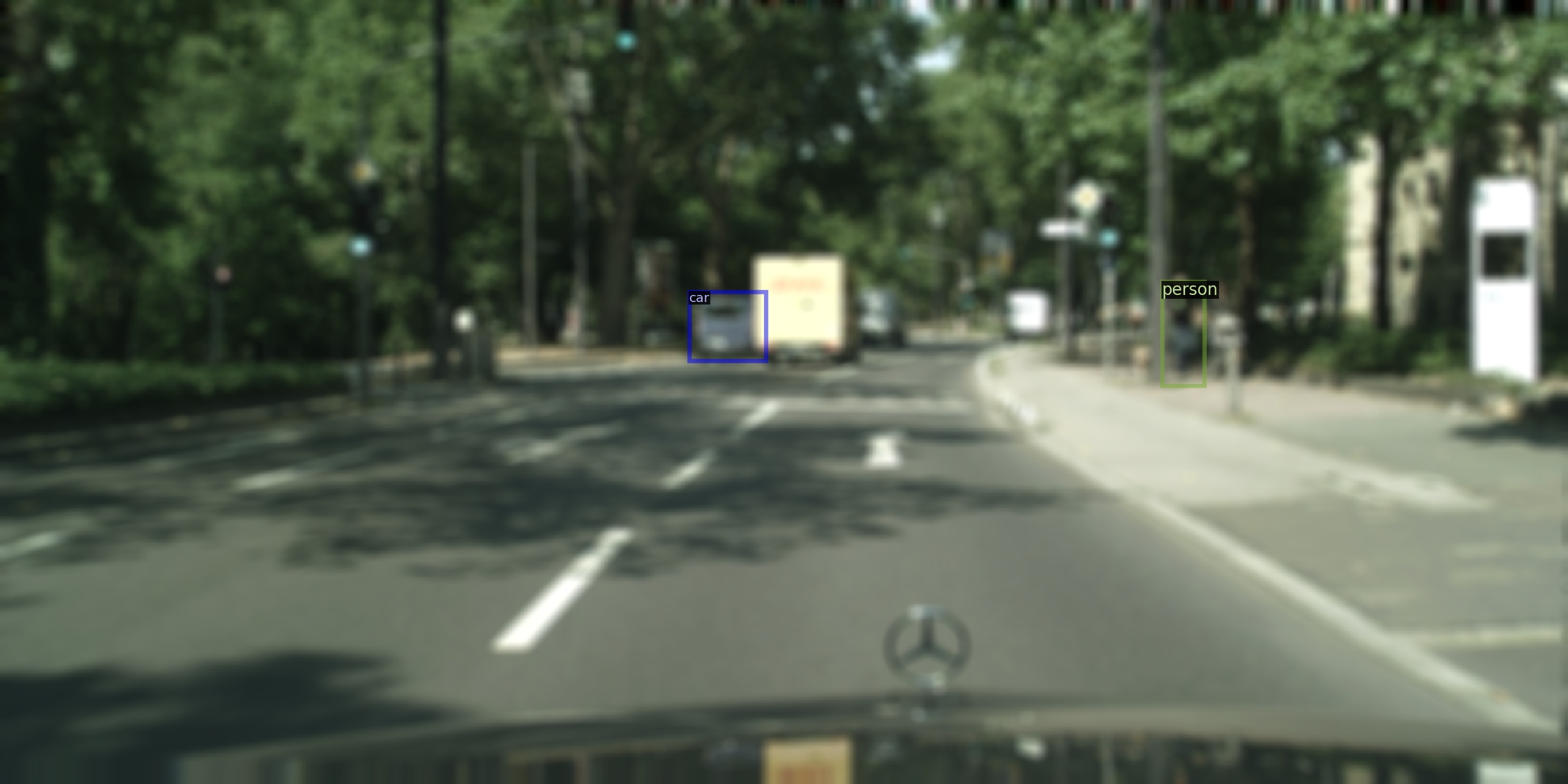}
      \label{dfig:subfig5}
  }
  \subfloat[AMROD (ours)]{
      \centering
      \includegraphics[width=0.32\textwidth]{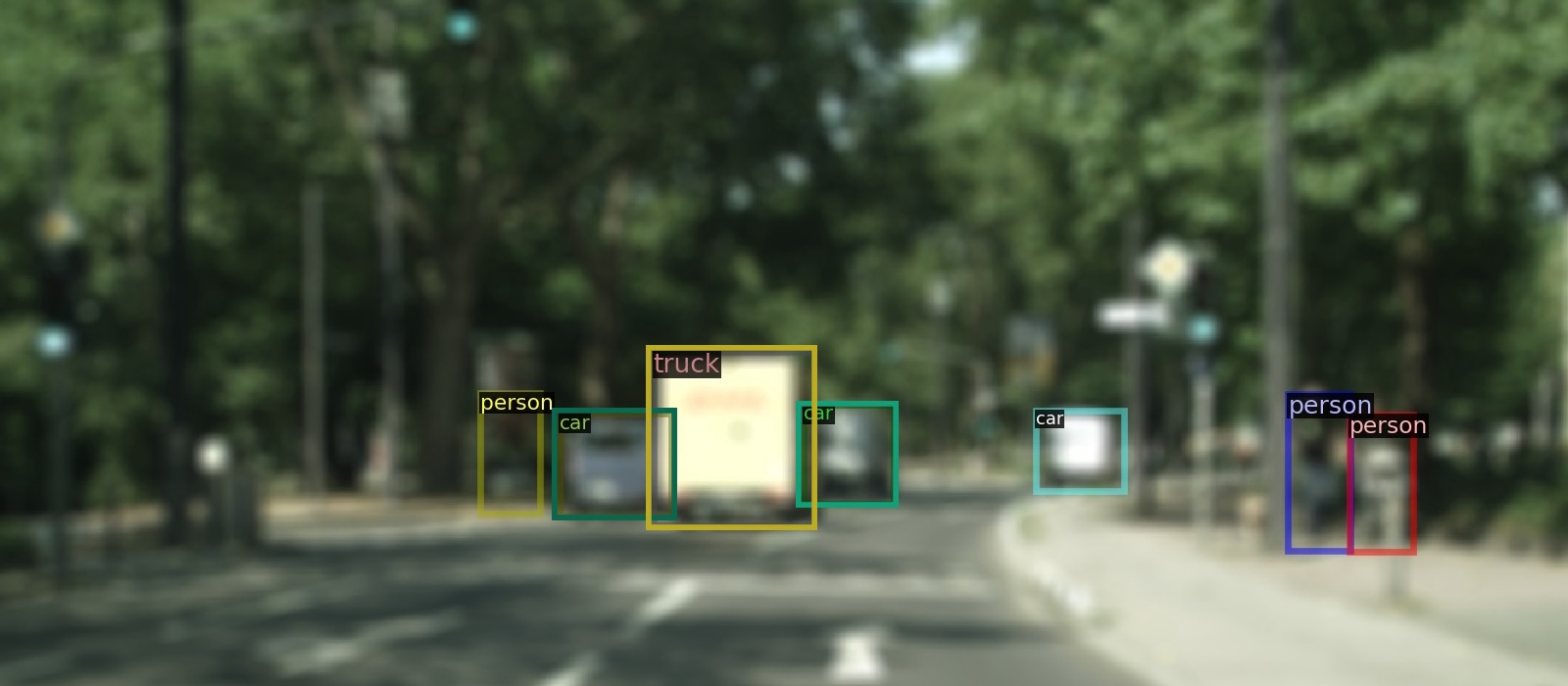}
      \label{dfig:subfig6}
  }
    \vspace{-10pt}
 \caption{Qualitative results. We compare the detection results of the AMROD and other baseline methods in the $10^{th}$ round of adaption to \textit{Defocus} corruption on the long-term Cityscapes-to-Cityscapes-C task.}
 \label{f2}
\end{figure}

As shown in Figure \ref{f1} and \ref{f2}, we compare the source model, baseline methods, and AMROD in the final adaptation to the brightness and defocus on the long-term Cityscapes-to-Cityscapes-C tasks.
AMROD guides the model in learning better feature representations while effectively mitigating forgetting in the long-term adaptation.
Therefore, AMROD assists detector in distinguishing more foreground object categories and better locating them.

\subsection{\textbf{Ablation Study}}
\subsubsection{Ablation Study Results on Model Components}

We conduct an ablation study to evaluate the impact of the OCL, AM, and ARR. 
As shown in Table \ref{ablation}, integrating the mean-teacher architecture achieves the performance of 25.9 mAP.
Building upon this foundation, the proposed OCL, AM, and ARR achieve an additional 1.5, 1.4, and 0.8 mAP improvement individually. 
Moreover, the combination of the OCL and AM modules raises the performance to 28.6 mAP.
Furthermore, including ARR mechanism brings a 0.6 mAP boost in performance, culminating in a model performance of 29.2 mAP.

\begin{table}[t]
\centering
\caption{Ablation experiment on Model Components. ``Mean-Teacher'' represents the base mean teacher framework with weak-strong augmentation and KL divergence distillation. 
All experiments are done on long-term Cityscapes-to-Cityscapes-C tasks.}
\resizebox{0.6\textwidth}{!}{
\begin{tabular}{cccc|cc}
\toprule
    Mean-Teacher & OCL & AM & ARR & Mean & Gain  \\ \midrule
  $\checkmark$ &    &     &     & 25.9     &  /     \\
 $\checkmark$ &  &  & $\checkmark$    & 26.7     & +0.8      \\
 $\checkmark$ &  & $\checkmark$ &     & 27.3     & +1.4      \\
 $\checkmark$ & $\checkmark$ &     &     & 27.4     & +1.5     \\
 $\checkmark$ & $\checkmark$ & $\checkmark$ &     & 28.6     & +2.7      \\
 \rowcolor{lightgreen}
 $\checkmark$ & $\checkmark$ & $\checkmark$ & $\checkmark$ & 29.2 & +3.3 \\ 
 \bottomrule
\end{tabular}
}
\label{ablation}
\end{table}

\begin{table}[tbp]
\centering
\caption{Ablation Study on Module Variants. The ``student'' variant utilizes predictions from the student model. The ``FT'' represents replacing AM module in AMROD with a fixed threshold. The ``SR'' and the ``DR'' indicate replacing ARR with the stochastic restoration \citep{wang2022continual} and the data-driven restoration \citep{brahma2023probabilistic}, respectively. 
 All experiments are done on long-term Cityscapes-to-Cityscapes-C tasks.}
\resizebox{0.7\textwidth}{!}{
\begin{tabular}{cccccll|l}
\toprule
   AMROD &Student & FT of 0.9 & FT of 0.8 & FT of 0.7 & SR & DR & Mean \\ 
\midrule
 $\checkmark$     & $\checkmark$     &                 &                        &                        &                        &                         & 27.9 \\
 $\checkmark$  &   & $\checkmark$                      &                        &                        &                        &                         & 27.7 \\
 $\checkmark$  &   &                        & $\checkmark$                      &                        &                        &                         & 28.2 \\
$\checkmark$   &  &                        &                        & $\checkmark$                      &                        &                         & 28.5 \\
 $\checkmark$  &   &                        &                        &                        & $\checkmark$                      &                         & 28.7 \\
 $\checkmark$  &   &                        &                        &                        &                        & $\checkmark$                       & 28.3 \\
\rowcolor{lightgreen}
 $\checkmark$   &  &                        &                        &                        &                        &                         & \textbf{29.2} \\ 
 \bottomrule
\end{tabular}
}
\label{alb}
\end{table}

\subsubsection{Ablation Study Results on Module Variants}

As presented in Table \ref{alb}, we further perform a fine-grained comparison study to assess the effectiveness of the modules of AMROD. 
This is done by replacing the teacher predictions with the student prediction in the Mean-Teacher framework \citep{tarvainen2017mean}, AM with a Fixed threshold (FT) strategy, and ARR with stochastic restoration (SR) \citep{wang2022continual} or data-driven restoration (DR) \citep{brahma2023probabilistic} mechanism. 
The findings indicate a variant utilizing student model predictions achieved only 27.9 mAP. 
This performance dip is likely because the student model, being actively updated, is less stable than the teacher model, which is updated via the EMA strategy to mitigate error accumulation.
Furthermore, replacing the AM module with FT strategies (FT of 0.9, 0.8, and 0.7) resulted in mAP scores of 27.7, 28.2, and 28.5, respectively.  
These are notably lower than AMROD's 29.2 mAP, underscoring the benefit of AM's adaptability to varying model confidence across diverse categories and domains, which FT inherently lack.
Similarly, when the ARR module is substituted with SR or DR, performance drop to 28.7 mAP and 28.3 mAP, respectively. 
This suggests that SR's randomness might discard valuable domain-specific knowledge, while DR could retain noisy parameters. 
In contrast, AMROD's ARR, which selectively resets inactive parameters with higher probability while integrating a stochastic component, better preserves essential knowledge and ensures robust adaptation.
\begin{figure}[tbp]
\centering
   \subfloat[Sensitivity to $\beta_t$]{
      \centering
      \includegraphics[width=0.33\textwidth]{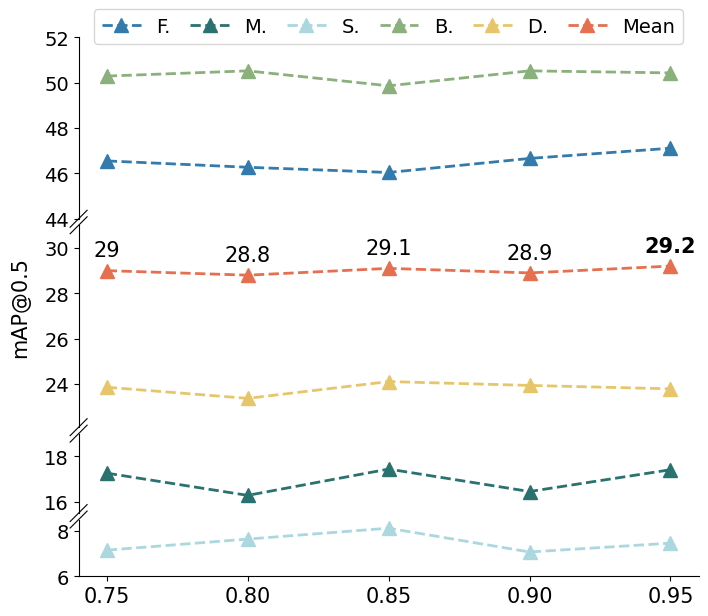} 
      \label{fig:subfig1}
  }
  \subfloat[Sensitivity to $\delta^0$]{
      \centering
      \includegraphics[width=0.33\textwidth]{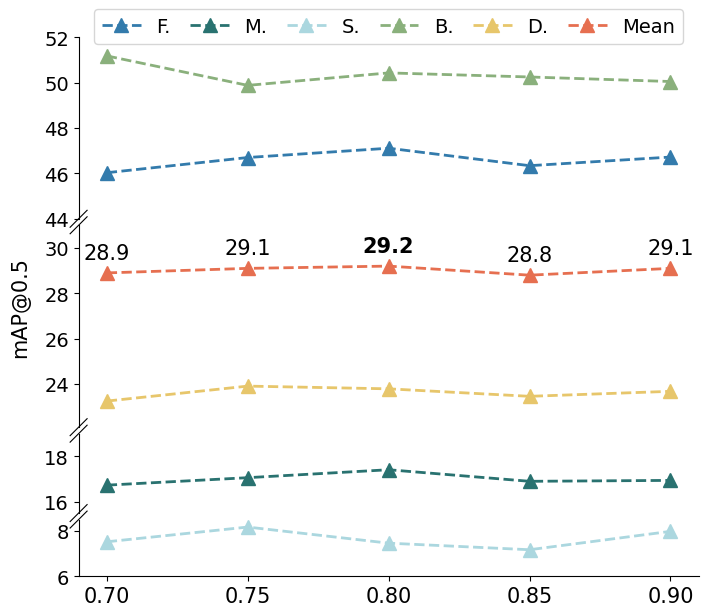}
      \label{fig:subfig2}
  }
  \subfloat[Sensitivity to $\epsilon$]{
      \centering
      \includegraphics[width=0.33\textwidth]{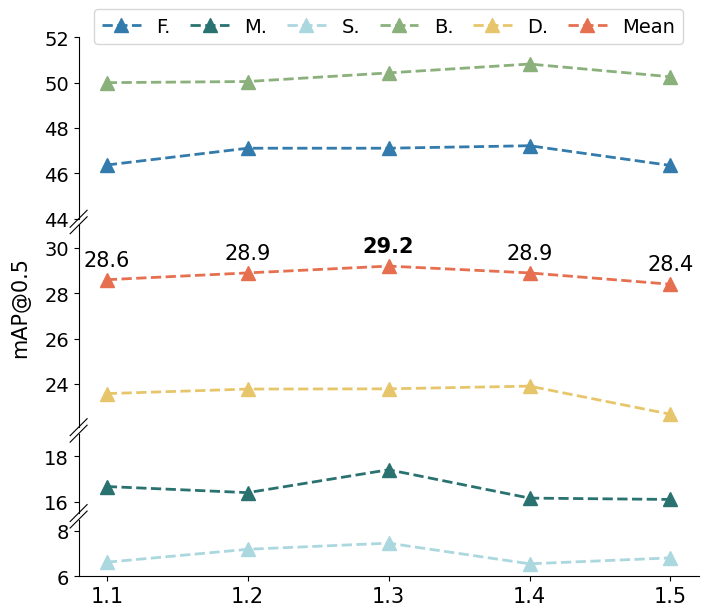}
      \label{fig:subfig3}
  }

  \subfloat[Sensitivity to $\beta_s$]{
      \centering
      \includegraphics[width=0.33\textwidth]{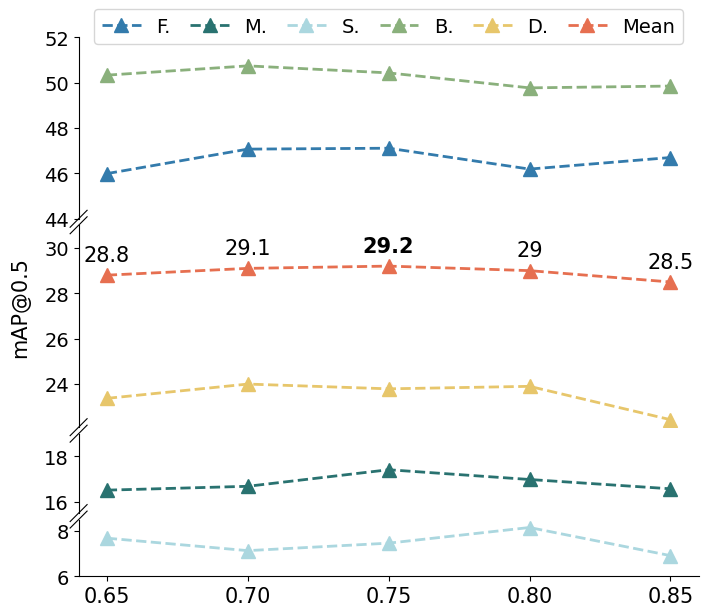} 
      \label{fig:subfig4}
  }  
  \subfloat[Sensitivity to $\delta_s$]{
      \centering
      \includegraphics[clip,width=0.33\textwidth]{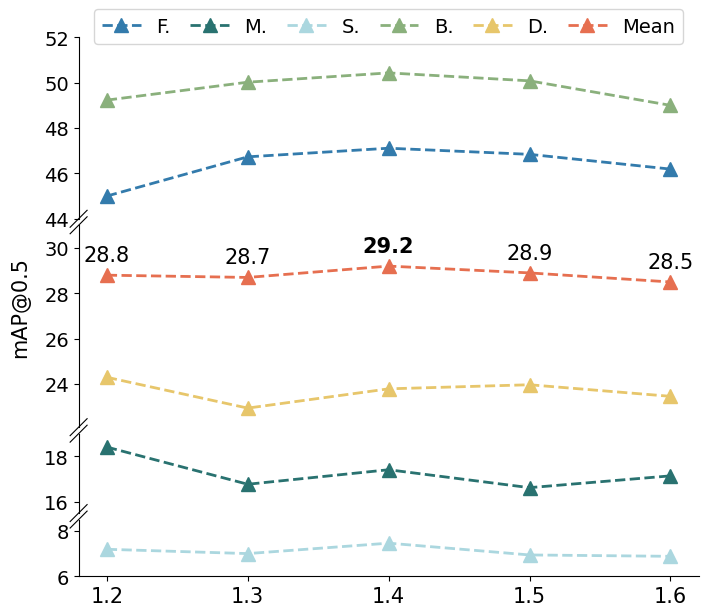}
      \label{fig:subfig5}
    
  }
  \subfloat[Sensitivity to $\eta$]{
      \centering
      \includegraphics[width=0.33\textwidth]{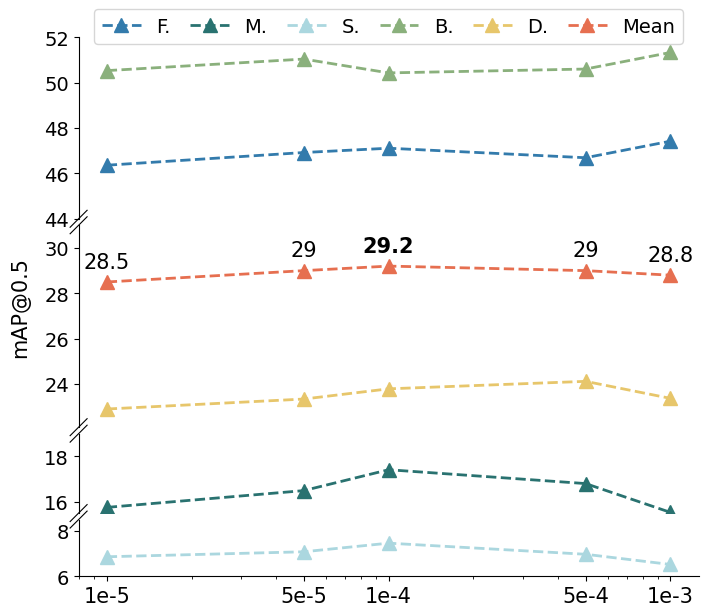}
      \label{fig:subfig6}
  }
  
 \caption{
Ablation Study on Hyperparameter Impact.
The hyperparameters are: $\beta_t$, $\delta^0$, and $\epsilon$ for dynamic threshold; $\beta_s$ and $\delta_s$ for dynamic skipping in AM; and $\eta$ for ARR.
The ``F.'', ``M.'', ``S.'', ``B.'', and ``D.'' represent the average performance over 10 times adaptation of Fog, Motion, Snow, Brightness, and Defocus, respectively.
All experiments are done on long-term Cityscapes-to-Cityscapes-C tasks.
}
 \label{hyper}
\end{figure}

\subsubsection{Ablation Study Results on Hyperparameter Impact}

To investigate the impact of hyperparameters, we conduct an ablation study on six key hyperparameters on the long-term Cityscapes-to-Cityscapes-C task, revealing the model's sensitivity and optimal configurations over the five target domains. 
These hyperparameters are varied while keeping others at their default settings, including $\beta_t$, $\delta^0$, $\epsilon$, $\beta_s$, $\delta_s$, and $\eta$.
As shown in Figure \ref{hyper}, for the AM module, the dynamic threshold update rate $\beta_t$ (Figure \ref{fig:subfig1}) shows peak performance at 0.95, suggesting that a balanced reliance on recent and historical category scores is beneficial.
The initial threshold $\delta^0$ (Figure \ref{fig:subfig2}) demonstrates strong robustness, while the linear projection factor $\epsilon$ (Figure \ref{fig:subfig3}) is optimal at 1.3, indicating that a moderate scaling of predicted scores best guides threshold adaptation.
For dynamic skipping, the update rate $\beta_s$ (Figure \ref{fig:subfig4}) achieves its best at 0.75 and the stability range parameter $\beta_s$ (Figure \ref{fig:subfig5}) performs best at 1.3, effectively balancing the reaction to mean score fluctuations for pausing adaptation.
Finally, the threshold $\eta$ (Figure \ref{fig:subfig6}) in ARR, which determines the q-quantile for resetting parameters, yields peak performance at 0.0001. 
Deviations from this value likely result in either insufficient resetting of inactive parameters or excessive resetting of potentially useful knowledge.
These findings highlight that the selected hyperparameters offer a balanced performance and confirm the stability of AMROD, as they demonstrate minimal sensitivity to small hyperparameter variations.

\subsection{Discussion}
Firstly, following prior works on TTA in objection detection \citep{vs2023towards,vs2023instance}, our experiments are primarily based on a single detector backbone, i.e., Faster R-CNN \citep{ren2015faster}. 
Additionally, AMROD is specifically designed for object detection tasks and has demonstrated its effectiveness exclusively in this domain.
Future work will extend to other advanced detector backbone networks or other vision tasks to assess the universality and generalization capabilities of AMROD.
Moreover, the proposed ARR mechanism to reset the model every iteration increases the computational cost. 
Future work could explore more efficient mechanisms for restoration, such as restoring the model only when significant changes in the distribution of the target domain are detected.
Finally, the evaluation tasks in our work are designed to simulate real-world adaptation scenarios by incorporating datasets affected by corruption and adverse conditions. 
However, real-world data distributions are inherently more complex. 
Consequently, a promising avenue for future research would be to apply our methodology in practical, real-world systems to further validate its effectiveness.

\section{Conclusion}
\label{s7}
In this work, we propose AMROD to address the two challenges in Continual Test-Time Adaptation (CTTA).
Firstly, object-level contrastive learning leverages ROI features for contrastive learning to refine the feature representation, tailored for object detection.
Secondly, the adaptive monitoring module enables efficient adaptation and high-quality pseudo-labels by dynamically skipping unstable adaptation and updating the category-specific threshold, based on the predicted confidence scores. 
Lastly, the adaptive randomized restoration mechanism selectively resets inactive parameters to mitigate forgetting while retaining valuable knowledge. 
{
Extensive empirical evaluations across four CTTA benchmarks demonstrate the effectiveness of AMROD for both short-term and long-term adaptation under synthetic and real-world continual distribution shifts. 
Notably, AMROD achieves state-of-the-art results on the long-term Cityscapes-to-Cityscapes-C task, improving accuracy by up to 3.2 mAP over existing baselines while simultaneously enhancing computational efficiency.
Ultimately, this work provides a robust, source-free framework capable of maintaining high-performance object detection in complex, non-stationary environments encountered by real-world perception systems.
}

\section*{Acknowledgments}
This work was supported by the Guangdong Science and Technology Program (2024B0101040005), National Natural Science Foundation of China (Grant No. T2125006 and No. 42401415), Shenzhen Science and Technology Program (KCXFZ20240903093759004 and KJZD20230923115106012), and Guangdong Science and Technology Program (2025B0101080001) (Corresponding author: Juepeng Zheng).

\bibliographystyle{elsarticle-harv} 
\bibliography{refs}

\end{document}